%% file: arxiv.tex
\documentclass{article}
\usepackage{iclr2027_conference,times}

\input{math_commands.tex}

\usepackage{url}

\usepackage{graphicx}
\usepackage{booktabs}
\usepackage{multirow}
\usepackage{wrapfig}
\usepackage[table]{xcolor}
\usepackage{makecell}    
\usepackage[most]{tcolorbox}
\usepackage{subcaption}
\usepackage{tabularx}
\usepackage{array}
\definecolor{headblue}{RGB}{178,223,219}
\definecolor{subpurple}{RGB}{225,210,238}
\definecolor{PromptTitleBlue}{RGB}{20, 40, 120}
\definecolor{PromptContentBlue}{RGB}{235, 240, 255}

\definecolor{mycitecolor}{RGB}{1,103,204}  % dark navy blue
\definecolor{mylinkcolor}{rgb}{0.55,0.15,0.15}
\usepackage[colorlinks=true,
            citecolor=mycitecolor,   % citations in dark navy blue
            linkcolor=mylinkcolor,         % keep cross-references black
            urlcolor=black]          % keep URLs black
           {hyperref}

\input{config}

\title{\method{}: Macro-Adaptive Parallel Reasoning for Regulating Search Momentum in Retrieval-Augmented Generation}

\iclrfinalcopy

\author{Xinze Li$^{1}$\thanks{\hspace{1mm}Equal Contribution.},\quad Dingwen Zhang$^{1}$\footnotemark[1],\quad Zhenghao Liu$^{1}$\thanks{\hspace{1mm}Corresponding Authors.},\quad Yukun Yan$^{2}$,\quad Shuo Wang$^{2}$,\quad  \\ \textbf{Yu Gu$^{1}$,\quad  Ge Yu$^{1}$,\quad Maosong Sun$^{2}$} \\
$^1$School of Computer Science and Engineering, Northeastern University, Shenyang, China \\
$^2$Department of Computer Science and Technology, Tsinghua University, Beijing, China\\}

\begin{document}

\maketitle
\lhead{Arxiv Preprint.}
% \begingroup
% \let\thefootnote\relax
% \footnotetext{$^{*}$ Equal contribution. $^{\dagger}$ Corresponding author.}
% \endgroup

\begin{abstract}
Retrieval-Augmented Generation (RAG) improves the performance of Large Language Models (LLMs) by retrieving external knowledge. Recently, some RAG methods interleave reasoning and retrieval, enabling LLMs to progressively generate sub-queries and retrieve external knowledge. However, these methods may suffer from redundant or ineffective sub-queries and insufficiently diverse query exploration. Moreover, due to the tightly interleaved nature of reasoning and retrieval, it remains unclear how the evolving reasoning process influences query intent. To characterize the evolution of query intent during reasoning, we introduce search momentum, which captures the accumulated retrieval tendency induced by historical changes in generated sub-queries along the reasoning trajectory. Through momentum-based analysis, we find that the query generation process tends to continuously evolve along existing retrieval trajectories. Based on this observation, we propose \method{}, a Macro-Adaptive Parallel Reasoning framework for regulating search momentum and mitigating query-intent drift. \method{} first retains the retrieved knowledge and a refined reasoning state, thereby reducing the influence of historical reasoning trajectories on subsequent sub-query generation. Subsequently, \method{} extends the conventional sequential reasoning process into a parallel reasoning paradigm. At each iteration, \method{} identifies complementary retrieval directions, generates corresponding sub-queries for parallel retrieval, and consolidates the retrieved evidence to update the knowledge and reasoning states. Experiments on six question-answering benchmarks demonstrate that \method{} consistently outperforms representative RAG methods across different backbone models. All code and resources are available at \url{https://github.com/NEUIR/MAPS}.
\end{abstract}

\input{sections/01_introduction}

\input{sections/02.5_prelimary}

\input{sections/03_methodology}

\input{sections/04_experiments}

\input{sections/05_results}
\input{sections/02_related_work}
\input{sections/06_conclusion}

\bibliographystyle{iclr2027_conference}
\bibliography{references}

\appendix
\input{sections/07_appendix}

\end{document}

%% file: math_commands.tex
\usepackage{amsmath,amsfonts,bm}

\def\eqref#1{equation~\ref{#1}}
\def\1{\bm{1}}

\DeclareMathAlphabet{\mathsfit}{\encodingdefault}{\sfdefault}{m}{sl}
\SetMathAlphabet{\mathsfit}{bold}{\encodingdefault}{\sfdefault}{bx}{n}

%% file: config.tex
\newcommand{\method}{MAPS}

\usepackage[most]{tcolorbox}
\usepackage{listings}
\newtcblisting{promptbox}[1]{
  enhanced,
  breakable,
  colback=white,
  colframe=gray!60,
  colbacktitle=gray!20,
  coltitle=black,
  fonttitle=\bfseries,
  title={#1},
  listing only,
  listing options={
    basicstyle=\ttfamily\scriptsize,
    breaklines=true,
    breakindent=0pt,
    columns=fullflexible,
    keepspaces=true,
    showstringspaces=false,
  },
  boxrule=0.5pt,
  left=5pt, right=5pt, top=5pt, bottom=5pt,
}

%% file: sections/01_introduction.tex
\section{Introduction}
\label{sec:intro}

Large Language Models (LLMs)~\citep{jaech2024openai,guo2025deepseek} have demonstrated remarkable capabilities in knowledge-intensive question answering and complex reasoning. However, they remain susceptible to hallucinations~\citep{jiang2023active,ji2023survey}, leading to factually incorrect or unsupported responses. Retrieval-Augmented Generation (RAG)~\citep{lewis2020retrieval,izacard2023atlas} addresses this issue by enabling LLMs to access external knowledge during inference, thereby improving the factuality and reliability of their responses. However, for multi-hop queries~\citep{yang2018hotpotqa}, the required evidence is often distributed across multiple documents and interconnected through dependencies, making single-step retrieval insufficient to comprehensively acquire the information needed to answer the query~\citep{fang2025kirag,wu2026finding}.

To address this limitation, recent RAG approaches have explored leveraging the reasoning capabilities of LLMs to adaptively acquire external evidence. These methods~\citep{li2025search,jin2025search,song2025r1} interleave reasoning and retrieval, allowing LLMs to identify missing information, generate sub-queries, and retrieve relevant documents to progressively support the original query. Specifically, an LLM is prompted to generate sub-queries when additional knowledge is necessary, retrieve relevant documents, and incorporate the retrieved evidence into subsequent reasoning. However, as reasoning and retrieval proceed iteratively, these methods can suffer from redundant or ineffective sub-queries, excessively long retrieval chains, or insufficient exploration diversity~\citep{ji2025evorag,martinez2025preqragclassifyrewrite}, ultimately degrading retrieval and downstream reasoning performance. Due to the tightly interleaved nature of reasoning and retrieval, it remains unclear how the evolving reasoning process influences the search direction, particularly when comprehensive evidence requires exploration beyond the current retrieval trajectory.

To characterize how search intentions evolve throughout the reasoning process, we introduce \emph{search momentum}, which captures the accumulated retrieval tendency induced by historical changes in generated sub-queries along the reasoning trajectory. At each step, we update the search momentum by aggregating the current sub-query with the accumulated momentum through a stepwise learnable weight. We learn these weights by using the updated momentum to predict the query-intent drift at the current step. By analyzing the resulting search momentum across reasoning steps, we observe that the deviation directions of subsequent sub-queries become increasingly aligned with the accumulated search momentum as retrieval proceeds deeper. This finding suggests that query generation tends to follow the existing retrieval trajectory, rather than correcting or escaping accumulated deviations through subsequent reasoning~\citep{guo2025deepseek}. Such momentum-driven behavior can progressively amplify early search deviations, causing the retrieval process to drift away from the original information need. This motivates us to regulate the accumulation of search momentum and preserve alignment between the evolving retrieval process and the original query intent.

In this paper, we propose \method{}, a Macro-Adaptive Parallel Reasoning framework for regulating search momentum and mitigating query-intent drift. \method{} introduces a parallel reasoning paradigm that reduces the risk of search momentum being dominated by a single deviated retrieval direction, thereby maintaining better alignment with the original query intent. Specifically, \method{} first disentangles reasoning traces from the sub-queries that propagate historical retrieval tendencies, retaining only the refined retrieval knowledge and reasoning state to reduce the influence of historical search deviations on subsequent query generation. It then extends the conventional sequential reasoning process into a parallel reasoning paradigm, enabling the model to calibrate the search direction and construct a more robust search momentum. Specifically, \method{} identifies multiple complementary retrieval directions from the current reasoning state and generates corresponding sub-queries in parallel to explore these directions. The resulting evidence is consolidated to update the accumulated knowledge and refined reasoning state, which serves as the basis for the next reasoning iteration.

Our experiments on multiple datasets demonstrate that \method{} achieves significant improvements over all baseline methods across different backbone models, highlighting its strong generalization ability and robustness. Further analyses show that \method{} effectively regulates the evolution of search momentum, preventing it from being dominated by a single retrieval direction. Our experiments further demonstrate that \method{} effectively mitigates query intent drift and maintains stronger relevance between generated sub-queries and supporting evidence throughout iterative reasoning. Moreover, \method{} enables the model to acquire the prerequisite knowledge required for reasoning at an earlier stage, while reducing unnecessary retrieval iterations.

%% file: sections/02.5_prelimary.tex
\section{Analyzing Query Intent Drift During LLM Reasoning}
\label{sec:2}
Recent RAG methods~\citep{li2025search} enable LLMs to actively retrieve relevant documents and incorporate them into the reasoning process, allowing them to progressively acquire the information required to answer a given query. However, the sub-queries generated during this iterative process do not necessarily remain aligned with the original information need. This is because subsequent sub-query generation depends on the preceding retrieval trajectory, which may contain sub-queries deviating from the original information need and noisy retrieved knowledge, thereby influencing the intent of subsequent sub-queries and ultimately degrading downstream reasoning performance~\citep{niu2026bound}. To characterize how the history retrieval trajectory influences the subsequent query intent, we first formalize the accumulated change in retrieval intent along the reasoning trajectory as \emph{search momentum}, and then investigate whether it can serve as a predictor of subsequent query intent drift.

\textbf{Uncovering Search Momentum in LLM Reasoning.} To characterize how retrieval intent evolves during iterative reasoning, we introduce \emph{search momentum}, which captures the accumulated semantic shift in search behavior across reasoning steps. Given an initial query $q_0$, suppose that the LLM has completed the reasoning trajectory up to the $(t-1)$-th step:
\begin{equation}
R_{t-1}
=
(r_1,q_1,o_1,\ldots,r_{t-1},q_{t-1},o_{t-1}),
\label{eq:trajectory}
\end{equation}
where $r_i$ denotes the reasoning step, $q_i$ is the generated sub-query, and $o_i$ denotes the knowledge observation obtained from the retrieved evidence. Conditioned on the initial query $q_0$ and the preceding reasoning trajectory $R_{t-1}$, the LLM $\mathcal{M}$ iteratively generates the next reasoning step and sub-query $q_t$ until reaching the maximum number of reasoning steps $T$:
\begin{equation}
r_t,q_t
=
\mathcal{M}(q_0,R_{t-1}).
\label{eq:subquery_generation}
\end{equation}
The generated sub-query $q_t$ is then used to retrieve a set of documents $D_t$, from which the LLM derives the corresponding observation $o_t$. The reasoning trajectory is subsequently updated as:
\begin{equation}
R_{t} = R_{t-1} \oplus (r_{t},q_{t},o_{t}).
\label{eq:trajectory_1}
\end{equation}

We represent the $t$-th step query $q_t$ by its semantic embedding $\bm{q}_t$ and characterize the evolution of retrieval intent through the semantic displacement between these sub-queries $\{q_0, q_1, \dots, q_T\}$:
\begin{equation}
\bm{q}_t=\mathrm{Emb}(q_t),
\end{equation}
where $\text{Emb}(\cdot)$ denotes the embedding model, which can be instantiated with the E5 model~\citep{wang2022e5}. We then define the search momentum $\mathbf{m}_t$ at the $t$-th reasoning step as:
\begin{equation}
\mathbf{m}_t
=
\beta_t\mathbf{m}_{t-1}
+
(1-\beta_t)(\bm{q}_t-\bm{q}_{t-1}),
\label{eq:search_momentum}
\end{equation}
where $\mathbf{m}_0=\mathbf{0}$ and $0\leq\beta_t<1$. The coefficient $\beta_t$ controls how strongly previously accumulated search dynamics are retained when incorporating the latest semantic shift. In this formulation, $\mathbf{m}_t$ is an exponentially weighted aggregation of historical query movements along the trajectory $q_0\rightarrow\cdots\rightarrow q_t$. Its magnitude reflects the strength of the accumulated semantic shift, while its direction characterizes the predominant direction in which the retrieval intent has evolved.

\input{table/momentum_beta_fit}
\textbf{Tracking Query Intent Drift with Search Momentum.}
After establishing search momentum as an explicit representation of accumulated query movement, we next examine whether it can predict the direction of subsequent query drift during LLM reasoning. Specifically, if $\mathbf{m}_t$ is consistently aligned with the deviation of the next sub-query from the original query, it suggests that historical search dynamics are propagated to subsequent query generation and that search momentum can serve as an effective signal for tracking intent drift.

For each reasoning trajectory, we therefore measure the directional alignment between the search momentum $\mathbf{m}_t$ and the deviation of the subsequent sub-query $q_{t+1}$ from the original query $q_0$:
\begin{equation}
S_t
=
\cos(\mathbf{m}_t,\bm{q}_{t+1}-\bm{q}_0),
\label{eq:momentum_alignment}
\end{equation}
where $\bm{q}_{t+1}-\bm{q}_0$ represents the semantic deviation of the next sub-query from the original information need. A larger $S_t$ indicates that the accumulated search dynamics captured by $\mathbf{m}_t$ are more consistent with the direction in which the subsequent query deviates from the original query, suggesting stronger predictive correspondence between search momentum and future query drift. 

To obtain the momentum at each reasoning step, we fit the coefficient $\beta_t$ in Eq.~\ref{eq:search_momentum} by maximizing the corresponding directional alignment $S_t$. Since $\beta_t$ is constrained to $[0,1)$, we solve the resulting bounded optimization problem using the L-BFGS-B quasi-Newton method~\citep{byrd1995limited}. Table~\ref{tab:momentum-beta-fit} reports the momentum coefficients fitted on the training set and the resulting directional alignment scores $S_t$ evaluated on a held-out set. The complete procedure is provided in Appendix~\ref{app:momentum-fitting}. We observe a consistent increase in both $\beta_t$ and $S_t$ as reasoning progresses. The increasing $\beta_t$ indicates that later reasoning steps place progressively greater weight on previously accumulated search dynamics, suggesting that historical queries exert an increasingly persistent influence on subsequent query generation as the retrieval process becomes deeper. Meanwhile, the increasing $S_t$ shows that the next query becomes more aligned with the current search momentum, suggesting that the reasoning process tends to perpetuate the existing search trajectory rather than effectively correcting or escaping it through reflection~\citep{guo2025deepseek}. This observation suggests that the reasoning process does not naturally eliminate the accumulated search momentum through subsequent reasoning steps. Taken together, these observations motivate us to investigate how to regulate the accumulation and propagation of search momentum, with the goal of reducing the persistent influence of historical search intents and maintaining better alignment with the original information need.

%% file: table/momentum_beta_fit.tex
% \begin{table*}[t]
%   \centering
%   \caption{Fitted Momentum Coefficients and Directional Alignment Between Search Momentum and Subsequent Query Intent Deviations. $S_t$ denotes the mean directional alignment score $S_t$ across trajectories in the independent test set. The Pooled MHQA row fits and evaluates on the union of the three datasets' respective training and test splits.}
%   \label{tab:momentum-beta-fit}
%   \resizebox{\textwidth}{!}{%
%     \begin{tabular}{lccccccccc}
%       \toprule
%       Dataset & $\beta_1$ & $S_1$ & $\beta_2$ & $S_2$ & $\beta_3$ & $S_3$ & $\beta_4$ & $S_4$ & Avg. \\
%       \midrule
%       2WikiQA     & 0.0000 & 0.4051 & 0.6352 & 0.5698 & 0.7552 & 0.6483 & 0.8761 & 0.6955 & 0.5797 \\
%       MuSiQue     & 0.0000 & 0.3105 & 0.5400 & 0.5196 & 0.7214 & 0.6489 & 0.8252 & 0.6913 & 0.5426 \\
%       HotpotQA    & 0.0000 & 0.4427 & 0.6161 & 0.6840 & 0.8151 & 0.7731 & 0.8836 & 0.7750 & 0.6687 \\
%       \midrule
%     Avg. & 0.0000 & 0.3875 & 0.5959 & 0.5651 & 0.7559 & 0.6774 & 0.8630 & 0.7279 & 0.5895 \\
%       \bottomrule
%     \end{tabular}%
%   }
% \end{table*}

\begin{table*}[t]
  \centering
  \caption{Fitted Momentum Coefficients and Directional Alignment Between Search Momentum and Subsequent Query Intent Deviations. $S_t$ denotes the mean directional alignment score $S_t$ across trajectories in the independent test set. $\beta_t$ controls the retention of previously accumulated retrieval tendencies when constructing search momentum. The Pooled-MHQA fits the momentum coefficients on the mixed training data from the three datasets and evaluates directional alignment on the corresponding test sets. Detailed experimental settings and results are provided in Appendix~\ref{app:momentum-fitting}.}
  \label{tab:momentum-beta-fit}
  \resizebox{\textwidth}{!}{%
    \begin{tabular}{lccccccccc}
      \toprule
      Dataset & $\beta_1$ & $S_1$ & $\beta_2$ & $S_2$ & $\beta_3$ & $S_3$ & $\beta_4$ & $S_4$ & Avg. $S_t$ \\
      \midrule
      2WikiQA     & 0.0000 & 0.4051 & 0.6352 & 0.5698 & 0.7552 & 0.6483 & 0.8761 & 0.6955 & 0.5797 \\
      MuSiQue     & 0.0000 & 0.3105 & 0.5400 & 0.5196 & 0.7214 & 0.6489 & 0.8252 & 0.6913 & 0.5426 \\
      HotpotQA    & 0.0000 & 0.4427 & 0.6161 & 0.6840 & 0.8151 & 0.7731 & 0.8836 & 0.7750 & 0.6687 \\
      \midrule
      Pooled-MHQA        & 0.0000 & 0.3875 & 0.5959 & 0.5651 & 0.7559 & 0.6774 & 0.8630 & 0.7279 & 0.5895 \\
      \bottomrule
    \end{tabular}%
  }
\end{table*}

%% file: sections/03_methodology.tex
\section{\method{}: Regulating Intent Drift through Parallel Reasoning}
\label{sec:method}
In this section, we propose \method{}, a Macro-Adaptive Parallel Reasoning-based RAG framework for regulating search momentum. \method{} introduces a parallel reasoning paradigm to reduce the influence of accumulated search momentum and maintain better alignment with the original query intent. Specifically, \method{} first disentangles historical reasoning traces and sub-queries that may propagate previously established retrieval tendencies, thereby reducing their influence on subsequent query generation. It then employs parallel reasoning to explore multiple complementary search directions, enabling the retrieval process to calibrate its search direction at an early stage and further mitigate query intent drift.

\textbf{Reasoning Disentanglement for Search Momentum Regulation.}
Given an initial query $q_0$, existing iterative RAG methods~\citep{li2025search} maintain a reasoning trajectory $R_{t-1}$ accumulated up to the $(t-1)$-th iteration:
\begin{equation}
R_{t-1}
=
(r_1,q_1,o_1,\ldots,r_{t-1},q_{t-1},o_{t-1}),
\end{equation}
where $r_i$, $q_i$, and $o_i$ denote the reasoning trace, generated sub-query, and knowledge observation obtained from retrieval at the $i$-th iteration, respectively. As discussed in Section~\ref{sec:2}, historical sub-queries can exert a stronger influence on subsequent query generation than the reasoning process itself, causing later queries to inherit and progressively reinforce previously established retrieval tendencies. Such trajectory-dependent propagation can consequently lead to query intent drift.

To mitigate this effect, \method{} disentangles the accumulated knowledge from the historical retrieval trajectory by not explicitly preserving the reasoning traces $r_i$ and sub-queries $q_i$ in $R_{t-1}$. Instead, it retains the knowledge acquired over the first $t-1$ iterations as a knowledge observation set:
\begin{equation}
O_{t-1}=(o_1,\ldots,o_{t-1}). 
\end{equation}
Meanwhile, rather than retaining the complete sequence of historical reasoning traces $r_{1:t-1}=(r_1,\ldots,r_{t-1})$, \method{} maintains a compact refined reasoning state $\tilde r_{t-1}$ that summarizes the essential reasoning information accumulated from previous iterations. This state is updated at each iteration while avoiding explicit exposure to the historical sequence of sub-queries.

Accordingly, at the $t$-th iteration, the subsequent reasoning generated by the model $\mathcal{M}_{\mathrm{Reason}}$ is conditioned on the original query $q_0$, the accumulated knowledge observations $O_{t-1}$, and the refined reasoning state $\tilde r_{t-1}$, rather than the complete historical trajectory $R_{t-1}$:
\begin{equation}\label{eq:reason_t}
(r_t, q_t, o_t) = \mathcal{M}_{\mathrm{Reason}} (q_0, O_{t-1}, \tilde r_{t-1}),
\end{equation}
where $\tilde r_{t-1}$ provides the reasoning information necessary for the current iteration, while $O_{t-1}$ preserves the knowledge acquired from previous retrievals. By removing the explicit dependency on historical sub-queries and reasoning traces, \method{} reduces the propagation of previously accumulated retrieval tendencies into subsequent query generation.

The reasoning process terminates when the accumulated knowledge $O_t=(o_1,\ldots,o_{t-1},o_t)$ satisfies the current information needs, or when the maximum number of iterations is reached. After the $T$-th iteration, the LLM generates the final answer $y$ based on the original query, the accumulated knowledge observations, and the refined reasoning state:
\begin{equation}
y=\mathcal{M}_{\mathrm{Reason}}(q_0,O_T,\tilde r_T),
\end{equation}
where $T$ denotes the termination iteration or the maximum number of allowed reasoning iterations.

\textbf{Parallel Reasoning for Search Direction Calibration.} To further regulate the search direction at the $t$-th iteration, \method{} introduces parallel reasoning to explore multiple complementary retrieval intents before generating the sub-queries. Specifically, we extend the sequential reasoning process in Eq.~\ref{eq:reason_t} to a parallel reasoning paradigm, where multiple retrieval directions are first identified and subsequently instantiated into independent sub-queries.

First, instead of directly generating a sub-query from the current reasoning state, we employ the search direction proposal model $\mathcal{M}_{\mathrm{Intent}}$ to identify multiple complementary retrieval directions:
\begin{equation}
\{s_t^1,\ldots,s_t^N\}
=
\mathcal{M}_{\mathrm{Intent}}
(q_0,O_{t-1},\tilde r_{t-1}),
\label{eq:direction}
\end{equation}
where each retrieval direction is instantiated as an individual sub-query. We then collect the generated sub-queries into a query set $Q_t=\{q_t^1,\ldots,q_t^N\}$. Unlike sequential query generation, in which each query is directly conditioned on the preceding retrieval trajectory, parallel query generation explores multiple retrieval directions from the same reasoning state. This expands the search space at an early stage and increases the likelihood of retrieving evidence that addresses the original information needs, thereby reducing the risk that the search momentum becomes dominated by a single deviated retrieval direction.

Next, each sub-query $q_t^i\in Q_t$ independently retrieves the top-$K$ documents from the document collection $\mathcal{D}$. The retrieved documents are then aggregated into a unified evidence set:
\begin{equation}
q_t^i
=
\mathcal{M}_{\mathrm{Query}}
(q_0,s_t^i,\tilde r_{t-1}),
\label{eq:parallel_subquery}
\end{equation}
where each retrieval direction is associated with a generated sub-query. Subsequently, \method{} collects the sub-queries generated from all retrieval directions to form the query set $Q_t=\{q_t^1,\ldots,q_t^N\}$. By generating multiple sub-queries in parallel, \method{} expands the search space at an early stage and increases the opportunity to generate sub-queries that retrieve useful evidence for the original query. Then, we use each sub-query $q_t^i$ independently retrieves the top-$K$ documents from the document collection $\mathcal{D}$ and collect all retrieved documents:
\begin{equation}
D_t=\bigcup_{q_t^i\in Q_t}
\operatorname{Top-}K(q_t^i,\mathcal{D}).
\label{eq:retrieval}
\end{equation}

Finally, the refinement agent $\mathcal{M}_{\mathrm{Refine}}$ consolidates the retrieved evidence from different retrieval directions into a knowledge observation $o_t$ and updates the refined reasoning state from $\tilde r_{t-1}$ to $\tilde r_t$:
\begin{equation}
o_t,\tilde r_t
=
\mathcal{M}_{\mathrm{Refine}}
(q_0,
\{s_t^1,\ldots,s_t^N\},
Q_t,
\mathcal{D}_t,O_{t-1},
\tilde r_{t-1}).
\label{eq:refine}
\end{equation}
The resulting knowledge observation $o_t$ and refined reasoning state $\tilde r_t$ are then regarded as the outputs of Eq.~\ref{eq:reason_t} to support subsequent reasoning iterations.

%% file: sections/04_experiments.tex
\section{Experimental Methodology}
\label{sec:experiments}

This section first describes datasets and baselines. Then, we introduce the implementation details of our experiments. 

\textbf{Datasets.} We evaluate different methods on three categories of benchmarks covering different task scenarios. For single-hop QA tasks, we use NQ~\citep{kwiatkowski2019nq} and PopQA~\citep{mallen2023popqa}. For multi-hop QA tasks, we evaluate \method{} on 2WikiQA~\citep{ho20202wiki}, HotpotQA~\citep{yang2018hotpotqa}, and MuSiQue~\citep{trivedi2022musique}. For reasoning QA tasks, following previous work~\citep{li2025search}, we evaluate on the Diamond subset of GPQA~\citep{rein2024gpqa}, which contains $198$ graduate-level science questions. Except for GPQA, we randomly sample $1,000$ questions from each dataset for evaluation. 

\textbf{Baselines.} We compare \method{} with multiple representative baselines. Vanilla LLM prompts the LLM to generate answers without external knowledge, and Vanilla RAG~\citep{lewis2020retrieval} utilizes the original query to retrieve documents and generates answers based on them. For adaptive RAG models, Self-RAG~\citep{asai2024self} prompts the LLM to determine whether retrieval is needed and whether retrieved documents are used during decoding. For reasoning-driven RAG models, IRCoT~\citep{trivedi2023ircot} interleaves reasoning with retrieval, where retrieved documents are incorporated into subsequent reasoning. Search-o1~\citep{li2025search} further extends this paradigm by introducing agentic query generation and the Reason-in-Documents module. For sub-query optimization RAG models, S3~\citep{jiang-etal-2025-s3} trains a lightweight model to generate more effective sub-queries. ParallelSearch~\citep{zhao2025parallel} iteratively decomposes the original query into multiple complementary sub-queries and retrieves them in parallel to explore diverse knowledge.

\textbf{Implementation Details.} We use Qwen3-32B~\citep{qwen3} and Qwen3.6-35B-A3B~\citep{qwen36_35b_a3b} as the backbone models. The retrieval corpus is the English Wikipedia December 2018 dump~\citep{karpukhin2020dpr}, and we employ E5-base-v2~\citep{wang2022e5} as the dense retriever with FAISS~\citep{douze2024faiss} for efficient similarity search over the pre-built index. Each retrieval step returns the top-5 most relevant passages, and the maximum number of execution iterations is set to 10. Following previous work~\citep{song2025r1,li2026skr}, we adopt Cover-EM as the primary evaluation metric.

%% file: sections/05_results.tex
\section{Evaluation Results}
\label{sec:results}

\subsection{Overall Performance}
\label{sec:performance}

\input{table/performance}

As shown in Table~\ref{tab:performance}, we compare \method{} with a range of representative baselines across different tasks. Overall, \method{} consistently outperforms all baselines, with improvements of 6.8\% and 5.8\% on Qwen3-32B and Qwen3.6-35B-A3B, demonstrating the effectiveness of our parallel reasoning paradigm for mitigating query intent drift. Notably, \method{} delivers consistent gains across different backbone models, indicating its robustness and generalizability across model architectures.

Compared with Naive-RAG, \method{} achieves substantial improvements, showing that reasoning-guided iterative retrieval can acquire more comprehensive knowledge than one-pass retrieval. More importantly, compared with iterative reasoning-based RAG methods such as Search-o1, \method{} improves performance by over 9\%. This result suggests that repeatedly extending the retrieval process along a single reasoning trajectory can reinforce previously explored retrieval directions, whereas parallel retrieval provides complementary search directions and reduces the risk of search momentum being dominated by a single trajectory.
In addition, \method{} consistently outperforms ParallelSearch across all datasets. This comparison shows that simply generating multiple sub-queries to broaden the search space is insufficient to fully mitigate the influence of accumulated retrieval tendencies on subsequent query generation. In contrast, \method{} explicitly explores complementary retrieval directions while retaining only the retrieved knowledge, rather than the full historical reasoning trajectory, as the context for subsequent retrieval. By decoupling subsequent query generation from the accumulated reasoning trajectory, \method{} effectively mitigates search intent drift and enables more diverse and targeted knowledge acquisition.

\subsection{Ablation Study}
\label{sec:ablation}
In this subsection, we conduct ablation studies to evaluate the effectiveness of different components in \method{}. In the experiments, we compare \method{} with three ablated models. \method{} w/o Retrieval Direction directly generates multiple sub-queries in parallel based on the accumulated knowledge and refined reasoning state, without first identifying multiple retrieval directions to guide sub-query generation. \method{} w/o Parallel Reasoning replaces the multi-directional parallel reasoning mechanism with sequential single-query generation and retrieval, which generates only one retrieval direction and its corresponding sub-query at each iteration. \method{} w/o Disentanglement does not disentangle the retrieval trajectory, but instead preserves the complete reasoning traces and sub-queries generated at each iteration for subsequent reasoning.

As shown in Table~\ref{tab:ablation}, \method{} consistently outperforms all ablated variants, demonstrating the effectiveness of the proposed framework. Specifically, compared with \method{} w/o Retrieval Direction, \method{} achieves over a 4\% improvement, indicating that explicitly identifying complementary retrieval directions helps guide parallel retrieval toward diverse information needs. \method{} also outperforms \method{} w/o Parallel Reasoning, demonstrating that the parallel reasoning mechanism expands the search space to acquire supporting evidence at early stages, thereby preventing the subsequent retrieval process from being dominated by a single retrieval direction and causing query intent drift. Moreover, comparing \method{} w/o Disentanglement with \method{}, we observe additional gains, suggesting that retaining the complete historical retrieval trajectory introduces stronger dependencies on previous retrieval tendencies, propagating accumulated deviations to subsequent query generation.

\input{table/ablation}

\subsection{Evolution of Search Momentum during Iterative Reasoning}
\label{sec:momentum-analysis}
\input{image/momentum}

As shown in Figure~\ref{fig:2.1}, we analyze the evolution of search momentum across different methods and investigate whether \method{} can effectively regulate its directional drift. In this experiment, we randomly sample 500 instances from all benchmark datasets. Detailed experimental settings and additional results are provided in Appendix~\ref{app:search-momentum}.

As shown in Figures~\ref{fig:2.1.1} and~\ref{fig:2.1.2}, we apply t-SNE to jointly project the embeddings of the original queries and their corresponding search momentum at different reasoning steps into a two-dimensional space. Compared with \method{}, Search-o1 exhibits a larger directional deviation of search momentum from the original query. This observation suggests that, compared with the reasoning-based RAG method, \method{} mitigates the tendency of search momentum to become dominated by a single retrieval direction through its parallel reasoning mechanism. Furthermore, \method{} w/o Disentanglement exhibits weaker alignment between search momentum and the original query than \method{}, suggesting that disentangling accumulated knowledge from the reasoning states and generated sub-queries in the historical reasoning trajectory can reduce the influence of historical retrieval tendencies on the evolution of search momentum.
Finally, for Figure~\ref{fig:2.1.3}, we compute the change in cosine similarity between the search momentum and the original query across consecutive iterations and accumulate these changes to quantify the directional deviation of search momentum. The results demonstrate the effectiveness of \method{} in calibrating the search direction through parallel retrieval and reasoning disentanglement during reasoning. As the reasoning process progresses, the search momentum of \method{} becomes more stable than that of the other methods, indicating that \method{} can better maintain coherent search directions established in the early reasoning steps.

\subsection{Analysis of \method{} Retrieval Behavior during Iterative Reasoning.}
\label{sec:alignment-analysis}
\input{image/alignment}
In this section, we further analyze the retrieval behavior of \method{} during iterative reasoning. We use a pooled set of all datasets with ground-truth documents.

As shown in Figure~\ref{fig:alignment}, we first analyze how the similarity between sub-queries and the original query, as well as the golden evidence, evolves across iterations. As shown in Figure~\ref{fig:alignment-2wiki-intent}, we compute the cosine similarity between the queries generated at different reasoning iterations and the original query. The result shows that, during the early stages of reasoning, the similarity between the generated sub-queries and the original query decreases rapidly for both Search-o1 and \method{}, reflecting the need to explore additional information beyond the original query. However, as reasoning progresses, the sub-queries generated by \method{} consistently maintain higher similarity with the original query than those generated by Search-o1 and gradually become more stable. This suggests that \method{} calibrates search momentum at the early stages through parallel reasoning and reasoning disentanglement, enabling subsequently generated sub-queries to remain more closely aligned with the information need expressed by the original query. As shown in Figure~\ref{fig:alignment-2wiki-evidence}, for each sample, we compute the similarity between the sub-query generated at each iteration and each of its golden evidence, and then average these similarity scores. The results show that the similarity between the sub-queries generated by \method{} and the ground-truth evidence decreases during the early stages, but subsequently recovers and stabilizes. In contrast, the similarity for Search-o1 continues to decline as reasoning progresses. This indicates that, after exploring different retrieval directions through parallel reasoning at the early stages, \method{} can progressively calibrate the query direction, enabling subsequent queries to more stably follow retrieval directions that are aligned with the target evidence.

\input{image/iterations-with-scores}

As shown in Figure~\ref{fig:iterations-with-scores}, we further analyze at which iteration different methods terminate during iterative reasoning and their corresponding answer performance. Figure~\ref{fig:iterations-2wiki} reports the proportion of samples terminating at each iteration for different methods. The results show that \method{} has a larger proportion of samples terminating at early iterations and also exhibits a higher proportion of samples requiring five or more retrieval rounds. This suggests that, for simple queries, \method{} can calibrate the query generation at early stages through parallel reasoning, enabling the model to rapidly acquire sufficient evidence with fewer iterations and terminate in a timely manner. For complex questions, \method{} regulates search momentum to reduce the accumulation of historical retrieval tendencies, allowing subsequent queries to continue exploring relevant knowledge. Figure~\ref{fig:iterations-hotpot} further compares the answer performance of models at different termination iterations. The results show that \method{} maintains strong answer performance across different termination steps. This suggests that its early termination typically occurs after sufficient evidence has been acquired, rather than resulting from erroneous stopping caused by retrieval drift. Meanwhile, \method{} continues to achieve higher performance at later termination steps, indicating that, after regulating search momentum, subsequent sub-queries consistently retrieve along directions aligned with the original information need.

%% file: table/performance.tex
\begin{table*}[t]
  \caption{
    Overall Performance of \method{} across All Benchmarks. The best results are in \textbf{bold} and the second-best are \underline{underlined}.
  }
  \label{tab:performance}
  \centering
    \begin{tabular}{lccccccc}
      \toprule
      Method           & NQ                                 & PopQA                             & 2WikiQA          & HotpotQA         & MuSiQue          & GPQA             &  Avg.                                                            \\
      \midrule
      \multicolumn{8}{c}{\textbf{Qwen3-32B}}                                                                                                                                                   \\
      \midrule
      Vanilla LLM & 25.0                               & 19.4                              & 27.3             & 21.3             & 7.1              & 44.4 & 24.1             \\
      % CoT              & 26.1                               & 10.8                              & 17.7             & 16.4             & 5.6              & \underline{31.5} & 18.0             \\
     Vanilla RAG        & 49.0                               & 52.1                              & 30.0             & 38.3             & 9.9              & 44.2             & 37.2             \\

      Self-RAG         & 39.2                               & 37.9                              & 29.8             & 31.5             & 17.2             & 45.1             & 33.4             \\
      
      IRCOT            & 38.2                               & 37.6                              & 38.8             & 33.2             & 22.3             & 54.2             & 37.4             \\
      
      Search-o1        & 53.1                               & 56.5                              & 43.7             & 40.0             & 23.7             & 46.4             & 43.9             \\
      S3               & \underline{59.5}                   & \underline{60.6}                  & 37.2             & 38.3             & 19.4             & \underline{60.7}             & 45.9             \\
      ParallelSearch   & 57.2                               & 60.4                              & \underline{53.6} & \underline{43.5} & \underline{30.9} & 49.8             & \underline{49.2} \\
      MAPS             & \textbf{63.4}                      & \textbf{64.8}                     & \textbf{59.4}    & \textbf{48.8}    & \textbf{37.6}    & \textbf{62.1}    & \textbf{56.0}    \\
      \midrule
      \multicolumn{8}{c}{\textbf{Qwen3.6-35B-A3B}}                                                                                                                                             \\
      \midrule
      Vanilla LLM & 39.4                               & 39.4                              & 42.1             & 34.8             & 19.4             & 76.8             & 42.0             \\
      Vanilla RAG        & 55.6                               & 61.4                              & 49.1             & 45.5             & 26.8             & 75.7             & 52.3             \\
      Self-RAG         & 45.5                               & 52.3                              & 44.8             & 39.7             & 27.3             & 67.6 & 46.2             \\
      
      IRCOT            & 49.0                               & 45.9                              & 45.8             & 43.2             & 28.6             & 64.7             & 46.2             \\
      
      Search-o1        & 53.0                               & \underline{62.0}                  & \underline{56.3} & 44.0             & 34.1             & 76.3             & 54.3             \\
      S3               & 57.4                               & 59.8                              & 34.5             & 41.9             & 18.9             & 73.7             & 47.7             \\
      ParallelSearch   & \underline{58.4}                   & 56.8                              & 55.9             & \underline{53.7} & \underline{41.5} & \underline{77.5}             & \underline{57.3} \\
      MAPS             & \textbf{64.1}                      & \textbf{69.3}                     & \textbf{65.8}    & \textbf{55.1}    & \textbf{45.2}    & \textbf{79.2}    & \textbf{63.1}    \\
      \bottomrule
    \end{tabular}
\end{table*}

\vspace{-1em}

%% file: table/ablation.tex
\begin{table*}[t]
  \caption{Ablation Study. The best results are in \textbf{bold} and the second-best are \underline{underlined}.}
  \label{tab:ablation}
  \centering
  \resizebox{\textwidth}{!}{%
    \begin{tabular}{lccccccc}
      \toprule
      Method                   & NQ               & PopQA            & 2WikiQA          & HotpotQA         & MuSiQue          & GPQA             & Avg.             \\
      \midrule
      \multicolumn{8}{c}{\textbf{Qwen3-32B}}                                                                                                                        \\
      \midrule
      \method{} & \textbf{63.4}    & \textbf{64.8}    & \textbf{59.4}    & \textbf{48.8}    & \textbf{37.6}    & \textbf{62.3}    & \textbf{56.1}    \\
      \quad  w/o Retrieval Direction & \underline{63.0} & \underline{64.3} & \underline{55.8} & 46.9 & \underline{35.6} & \underline{46.0} & \underline{51.9} \\
    
      %\quad MAPS w/o Experience     & 59.3             & 63.2             & 54.9             & 46.8             & 34.3             & 41.1             & 49.9             \\
      \quad  w/o Parallel Reasoning & 60.1             & 61.2             & 54.1             & 45.2             & 33.1             & 45.3             & 49.8             \\
    \quad w/o Disentanglement & 61.0          & 58.8          & 53.1             & \underline{48.3}             & 34.3            & 45.1          & 50.1          \\
      \midrule
      \multicolumn{8}{c}{\textbf{Qwen3.6-35B-A3B}}                                                                                                               \\
      \midrule
      \method{} & \textbf{64.1}    & \textbf{69.3}    & \textbf{65.8}    & \textbf{55.1}    & \textbf{45.2}    & \textbf{79.2}    & \textbf{63.1}    \\
      \quad  w/o Retrieval Direction & \underline{62.3} & \underline{69.2} & 62.2             & 51.6 & \underline{43.1}             & 68.1             & 59.4             \\
     
      %\quad MAPS w/o Experience     & 58.2             & 66.5             & 60.5             & 50.8             & 42.5             & 78.3             & 59.5             \\
      \quad w/o Parallel Reasoning & 56.3             & 66.8             & 62.8 & 49.8             & 39.5             & 63.8             & 56.5             \\

       \quad  w/o Disentanglement & 54.0          & 64.9          & \underline{64.4}             & \underline{52.9}            & 42.6            & \underline{78.1}          & \underline{59.5}          \\
      \bottomrule
    \end{tabular}
  }
\end{table*}
\vspace{-1em}

%% file: image/momentum.tex
% \begin{figure}[bt]
% \centering
% \subfigure[Search Momentum of \method{} and Search-o1.] {
% \label{fig:2.1.1}
% \includegraphics[width=0.33\linewidth]{fig/orbit_contour/3d/orbit_contour_m_s.pdf}}
% \hfill
% \subfigure[Search Momentum of \method{} and \method{} w/o Disentanglement.] {
% \label{fig:2.1.2}
% \includegraphics[width=0.33\linewidth]{fig/orbit_contour/3d/orbit_contour_m_p.pdf}}
% \hfill
% \subfigure[Cumulative Directional Offset of Search Momentum.] {
% \label{fig:2.1.3}
% \includegraphics[width=0.29\linewidth]{new_fig/section-2/delta.pdf}}
% \caption{Evolution of Search Momentum during Iterative Reasoning. Figures~\ref{fig:2.1.1} and~\ref{fig:2.1.2} visualize the trajectories of search momentum through t-SNE. Figure~\ref{fig:2.1.3} shows the cumulative directional offset of search momentum with respect to the original query across reasoning iterations.}
% \label{fig:2.1}
% \vspace{-1em}
% \end{figure}

\begin{figure}[bt]
\centering

\begin{subfigure}[t]{0.33\linewidth}
\centering
\includegraphics[width=\linewidth]{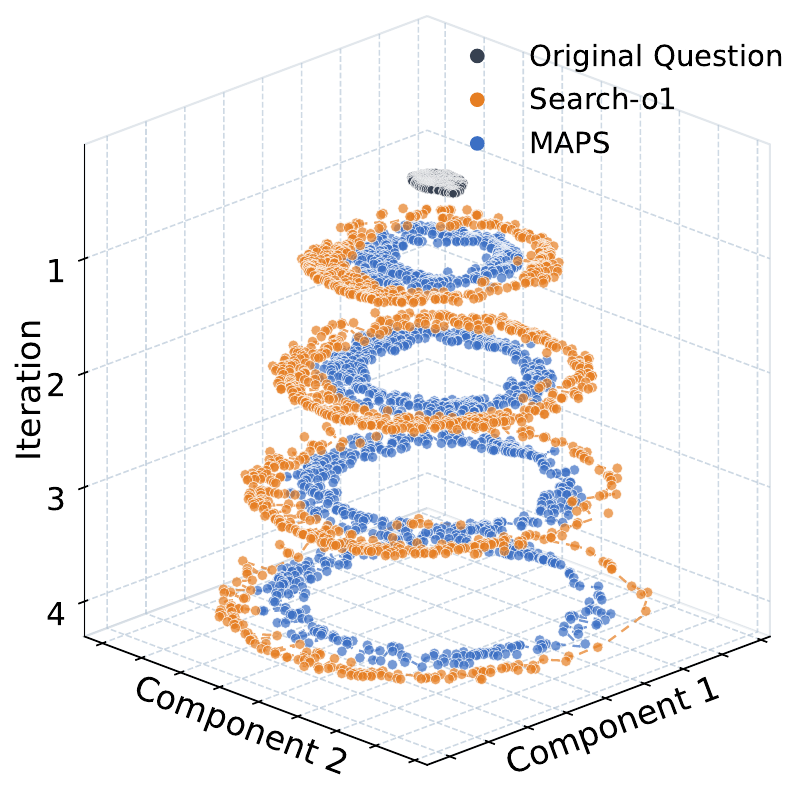}
\caption{Search Momentum of \method{} and Search-o1.}
\label{fig:2.1.1}
\end{subfigure}
\hfill
\begin{subfigure}[t]{0.33\linewidth}
\centering
\includegraphics[width=\linewidth]{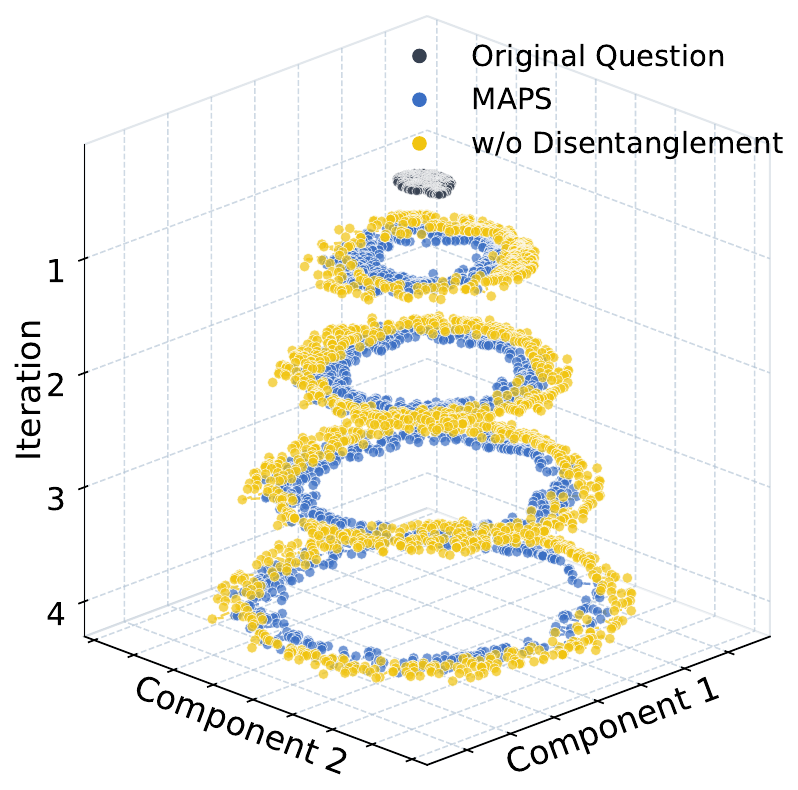}
\caption{Search Momentum of \method{} and \method{} w/o Disentanglement.}
\label{fig:2.1.2}
\end{subfigure}
\hfill
\begin{subfigure}[t]{0.29\linewidth}
\centering
\includegraphics[width=\linewidth]{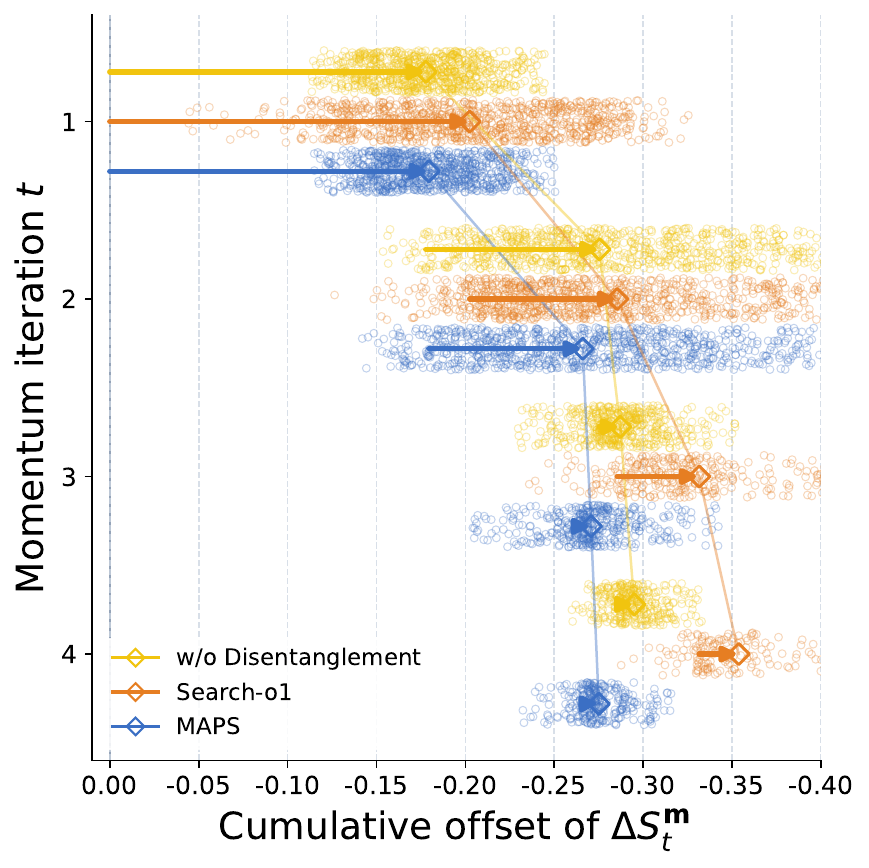}
\caption{Cumulative Directional Offset of Search Momentum.}
\label{fig:2.1.3}
\end{subfigure}

\caption{Evolution of Search Momentum during Iterative Reasoning. Figures~\ref{fig:2.1.1} and~\ref{fig:2.1.2} visualize the trajectories of search momentum through t-SNE. Figure~\ref{fig:2.1.3} shows the cumulative directional offset of search momentum with respect to the original query across reasoning iterations.}
\label{fig:2.1}
\vspace{-1em}
\end{figure}

%% file: image/alignment.tex
% \begin{wrapfigure}{r}{0.48\columnwidth}
%   \centering
%   \vspace{-1em}
%   \renewcommand{\theHsubfigure}{alignment.\arabic{subfigure}}
%   \setcounter{subfigure}{0}
%   \subfigure[Similarity between Queries.]{
%     \label{fig:alignment-2wiki-intent}
%     \includegraphics[width=0.46\linewidth]{fig/query-question-similiarty-heat/group/cosine/ms0_ts0_iteration/group_combined_MAPS-lower_Search-o1-upper_heat.pdf}}\hfill
%   \subfigure[Similarity between Queries and Golden Evidence.]{
%     \label{fig:alignment-2wiki-evidence}
%     \includegraphics[width=0.5\linewidth]{fig/query-gt-docs-similarity-line-gtsorted/group/cosine/group_per_gt_mean_full_text_ms0.pdf}}
%   \caption{Evolution of Query Similarity and Query-Evidence Similarity during Iterative Reasoning.}
%   \label{fig:alignment}
%   \vspace{-0.5em}
% \end{wrapfigure}

\begin{wrapfigure}{r}{0.48\columnwidth}
  \centering
  \vspace{-1em}

  \renewcommand{\theHsubfigure}{alignment.\arabic{subfigure}}
  \setcounter{subfigure}{0}

  \begin{subfigure}[t]{0.46\linewidth}
    \centering
    \includegraphics[width=\linewidth]{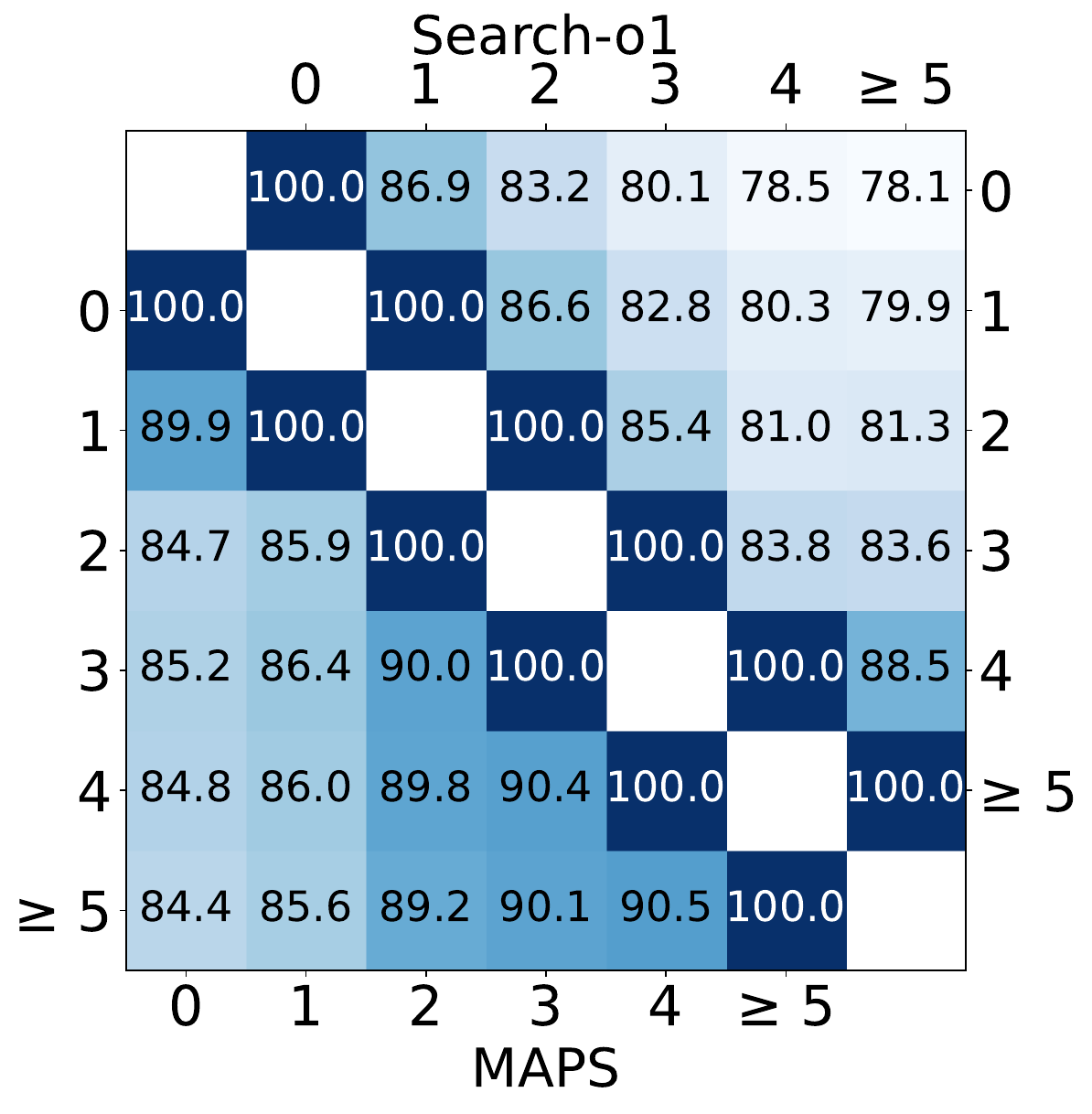}
    \caption{Similarity between Queries.}
    \label{fig:alignment-2wiki-intent}
  \end{subfigure}\hfill
  \begin{subfigure}[t]{0.5\linewidth}
    \centering
    \includegraphics[width=\linewidth]{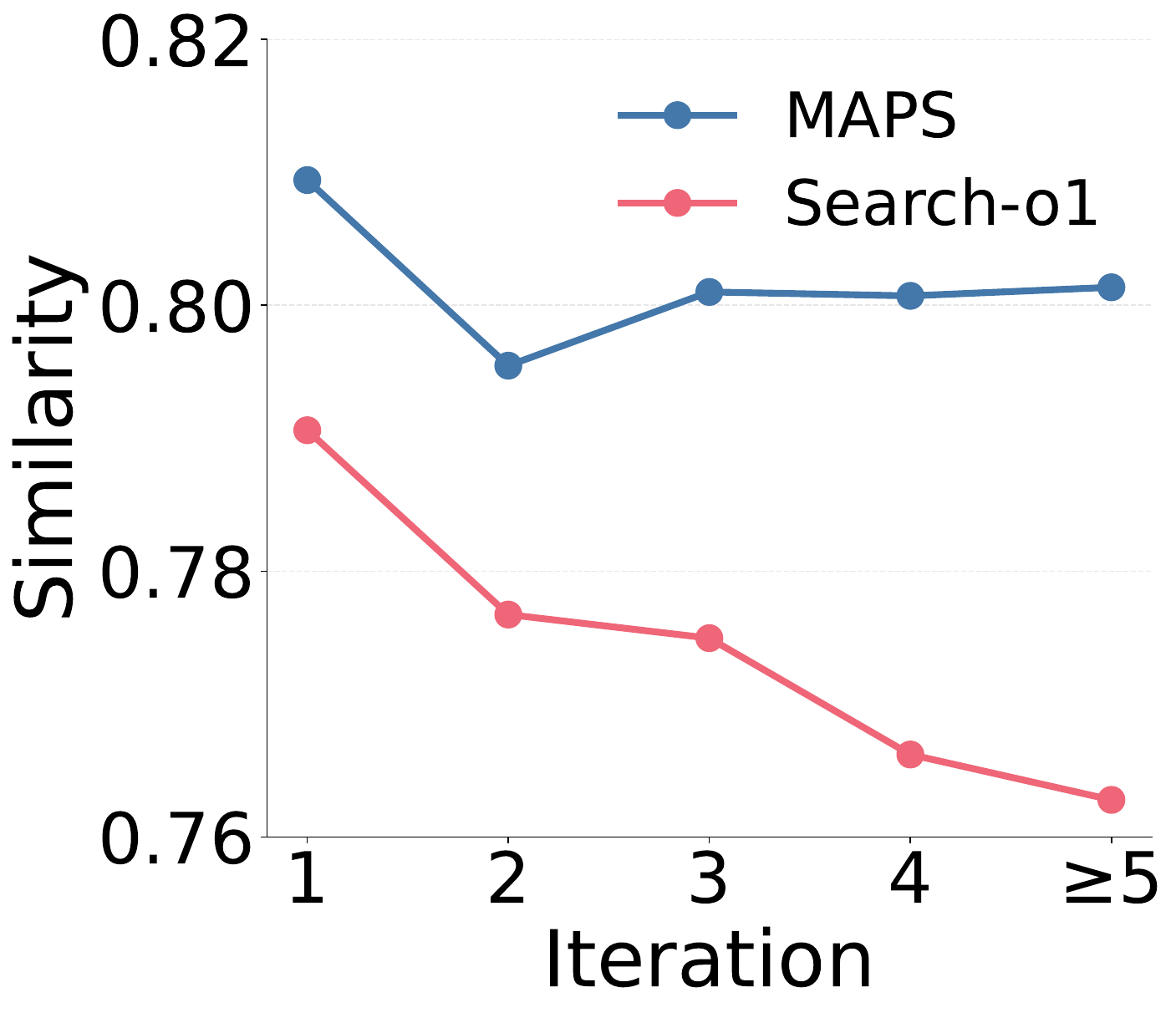}
    \caption{Similarity between Queries and Golden Evidence.}
    \label{fig:alignment-2wiki-evidence}
  \end{subfigure}

  \caption{Evolution of Query Similarity and Query-Evidence Similarity during Iterative Reasoning.}
  \label{fig:alignment}
  \vspace{-0.5em}
\end{wrapfigure}

%% file: image/iterations-with-scores.tex
% \begin{wrapfigure}{r}{0.48\columnwidth}
%   \centering
%   \vspace{-1em}

%   \renewcommand{\theHsubfigure}{iterations.\arabic{subfigure}}
%   \setcounter{subfigure}{0}
%    %\vspace{-1em}
%   \subfigure[Termination Ratio at Each Iteration.]{
%     \label{fig:iterations-2wiki}
%     \includegraphics[width=0.5\linewidth]{fig/iteration-termination/iteration-termination.pdf}}\hfill
%   \subfigure[Performance by Termination Step.]{
%     \label{fig:iterations-hotpot}
%     \includegraphics[width=0.4\linewidth]{fig/iterations-with-scores/group/group-combined-cover_em_1.pdf}}
%   \caption{Termination Distribution and Answer Performance across Different Methods.}

%   \label{fig:iterations-with-scores}
%   \vspace{-0.5em}
% \end{wrapfigure}

\begin{wrapfigure}{r}{0.48\columnwidth}
  \centering
  \vspace{-1em}

  \renewcommand{\theHsubfigure}{iterations.\arabic{subfigure}}
  \setcounter{subfigure}{0}

  \begin{subfigure}[t]{0.5\linewidth}
    \centering
    \includegraphics[width=\linewidth]{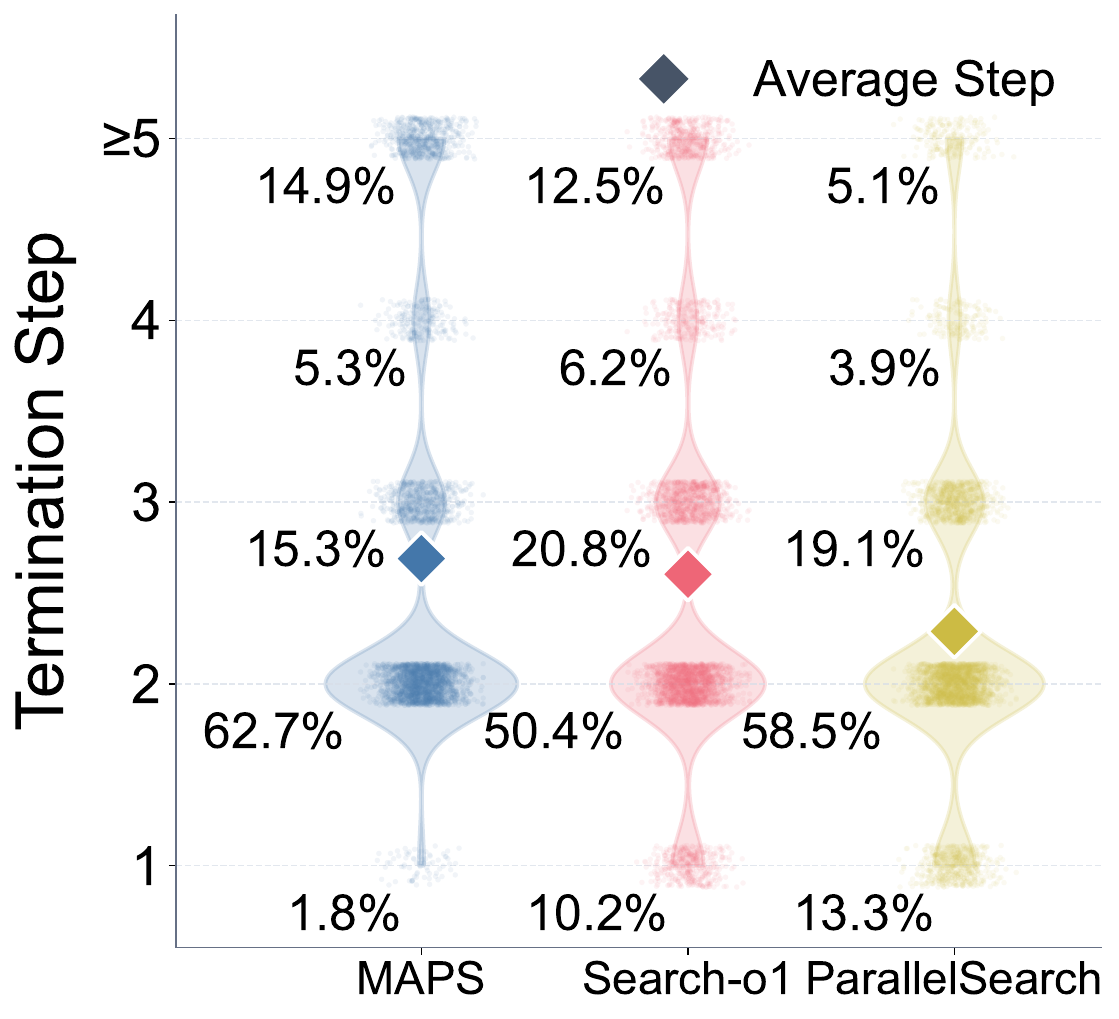}
    \caption{Termination Ratio at Each Iteration.}
    \label{fig:iterations-2wiki}
  \end{subfigure}\hfill%
  \begin{subfigure}[t]{0.45\linewidth}
    \centering
    \includegraphics[width=\linewidth]{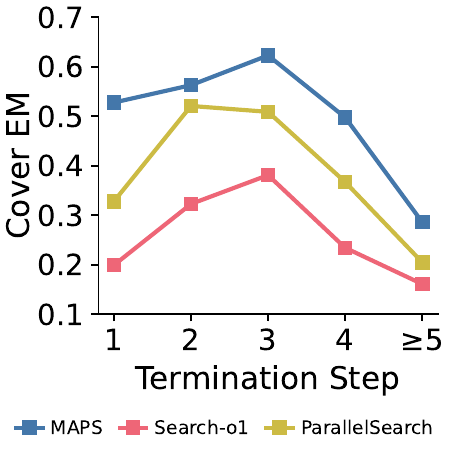}
    \caption{Performance by Termination Step.}
    \label{fig:iterations-hotpot}
  \end{subfigure}

  \caption{Termination Distribution and Answer Performance across Different Methods.}
  \label{fig:iterations-with-scores}
  \vspace{-0.5em}
\end{wrapfigure}

%% file: sections/02_related_work.tex
\section{Related Work}
\label{sec:related_work}

Retrieval-Augmented Generation (RAG) enhances Large Language Models (LLMs) by retrieving external knowledge during inference~\citep{lewis2020retrieval,izacard2023atlas,li2025rag}, thereby providing additional factual support and mitigating hallucinations~\citep{ji2023survey}. Early RAG systems typically follow a retrieve-then-read paradigm, where the model first retrieves relevant documents based on the original query and then feeds the retrieved documents as contextual input to prompt the LLM to generate responses~\citep{shi2024replug,izacard2023atlas,lin2024ra}. Furthermore, some studies investigate how to enable LLMs to adaptively determine when to retrieve external knowledge during response generation, thereby improving the reliability of generated responses~\citep{asai2024self}. However, for complex questions, retrieving solely with the original query is often insufficient to acquire all relevant evidence, as the required information is typically distributed across multiple documents~\citep{fang2025kirag,wu2026finding}.

Recent reasoning-driven RAG methods leverage the reasoning capabilities of LLMs to enable progressive reasoning and sub-query generation for acquiring external knowledge, while incorporating the retrieved evidence into subsequent reasoning processes~\citep{li2025search,jin2025search}. However, these methods heavily rely on the quality of generated sub-queries, where irrelevant or erroneous sub-queries may lead to deviations in subsequent retrieval trajectories. Therefore, some studies have begun to investigate how to optimize the sub-query generation process. S3~\citep{jiang-etal-2025-s3} trains a lightweight model specifically for sub-query generation, enabling the retrieval of more accurate knowledge. ParallelSearch~\citep{zhao2025parallel} prompts the model to generate multiple sub-queries at each iteration and performs parallel retrieval, thereby acquiring more comprehensive evidence. However, these methods do not eliminate the influence of historical retrieval trajectories on subsequent sub-query generation.

%% file: sections/06_conclusion.tex
\section{Conclusion}
\label{sec:conclusion}
In this paper, we introduce search momentum, which captures the accumulated retrieval tendency induced by historical changes in generated sub-queries along the reasoning trajectory. Our experiments show that the sub-queries generated by the model tend to continuously evolve along the accumulated retrieval trajectory, leading to deviations in query intent. Based on this observation, we propose \method{}, a Macro-Adaptive Parallel Reasoning framework for regulating search momentum and mitigating query-intent drift. By disentangling historical reasoning trajectories and introducing a parallel reasoning mechanism, \method{} effectively alleviates query-intent drift.

\section*{AI Use Statement}
Generative AI tools are used in this paper solely to check grammatical errors and assist with language polishing, with the aim of improving the clarity of the manuscript. All research ideas, experimental design, implementation, analysis, and conclusions are independently developed and verified by the authors without the use of any AI tools.

% Experiments on all benchmarks with different backbone models demonstrate that \method{} consistently outperforms representative RAG methods.

%% file: sections/07_appendix.tex
\newpage
\section{Appendix}
%Additional details and experimental results can be found in the appendix.

\subsection{Fitting the Search Momentum Coefficients}
\label{app:momentum-fitting}

In this section, we provide a detailed description of the experimental settings and methodology for fitting the momentum coefficients introduced in Section~\ref{sec:2}. Specifically, we first describe how the datasets used for momentum fitting are constructe. Then, we present the detailed procedure for estimating the search momentum coefficients. Finally, we report the fitted momentum results of the \method{} model.

\textbf{Data Construction for Momentum Fitting.} We collect 1,000 questions from each of 2WikiQA, MuSiQue, and HotpotQA to construct the training data as $\mathcal{D}_{\mathrm{train}}$ for fitting the momentum coefficients $\beta_t$. For each question, we run Search-o1 with Qwen3-32B as the backbone model and record the complete trajectory of iterative reasoning and retrieval. We discard trajectories with fewer than three retrieval iterations to retain those containing sufficient query transitions for analyzing accumulated search momentum. For each retained trajectory, we encode the original question and all generated sub-queries, $\{q_0,\ldots,q_T\}$, using the E5 embedding model, obtaining their semantic representations $\{\bm{q}_0,\ldots,\bm{q}_T\}$. To evaluate the directional alignment between search momentum and subsequent query deviations, we collect an additional 1,000 questions from each dataset  as $\mathcal{D}_{\mathrm{test}}$ , with no overlap with $\mathcal{D}_{train}$, and construct the held-out test data using the same filtering, and encoding procedures as train data do. The search momentum coefficients are fitted exclusively on the training data  $\mathcal{D}_{\mathrm{train}}$ and then fixed for evaluation on the test data  $\mathcal{D}_{\mathrm{test}}$. For the Pooled-MHQA setting, we merge the training data from the three datasets into a pooled training set and similarly combine their test data into a pooled test set.

\paragraph{Fitting Objective.} For each trajectory in the training set $\mathcal{D}_{\mathrm{train}}$, we first compute the stepwise directional alignment scores $\{S_1,\ldots,S_{T-1}\}$ according to Eq.~\ref{eq:momentum_alignment} in section~\ref{sec:2}. We then retain the first $K$ alignment scores from each trajectory, where $K=4$, since the vast majority of retrieval trajectories terminate within five iterations. Including later steps would substantially reduce the number of available trajectories at those positions and may lead to unreliable coefficient estimation due to the limited amount of training data.

The overall fitting objective averages the stepwise alignment scores equally:
\begin{equation}
J(\boldsymbol{\beta})
=
\frac{1}{K}
\sum_{i\in D_{train}}
\sum_{t=1}^{K}
S^{(i)}_t({\beta}).
\label{eq:app-fit-objective}
\end{equation}

For each trajectory $i$, we compute the momentum from Eq.~\ref{eq:search_momentum} with $\mathbf{m}_0=\mathbf{0}$ and use Eq.~\ref{eq:momentum_alignment} to obtain its stepwise alignment score $S_t^{(i)}(\boldsymbol{\beta})$. For \method{} and \method{} w/o Disentanglement, which use multiple sub-queries for retrieval at each iteration, we average the embeddings of all sub-queries generated at iteration $t$ and use the resulting vector as $\bm{q}_t$.
we estimate the search momentum coefficient $\boldsymbol{\beta}$ by minimizing the fitting objective $J(\boldsymbol{\beta})$ using the L-BFGS-B quasi-Newton method with 500 iterations.

\input{table/momentum_beta_fit_maps}
\input{table/momentum_beta_fit_maps_wo_disentanglement}

\paragraph{Evaluation and Results across Methods.}
After fitting, we freeze ${\boldsymbol{\beta}}$ and compute the held-out $S_t$ over eligible trajectories in $\mathcal{D}_{\mathrm{test}}$. We also report $\mathrm{Avg.}=\frac{1}{K}\sum_{t=1}^{K}S_t$.
%The first-step score $C_1$ is independent of the fitted free coefficients and serves as an initial alignment reference.
%We fit a separate coefficient vector for each of the three multi-hop datasets and Pooled MHQA under each method. 
Table~\ref{tab:momentum-beta-fit} reports the Search-o1 results in the main text; Tables~\ref{tab:momentum-beta-fit-maps} and~\ref{tab:momentum-beta-fit-maps-ablation} report the corresponding results for \method{} and \method{} w/o Disentanglement. Each table gives the fitted coefficients and held-out alignment scores for all four dataset settings.

\subsection{Experimental Details for Analyzing the Evolution of Search Momentum}
\label{app:search-momentum}
In this section, we describe how the three-dimensional search momentum visualizations in Figures~\ref{fig:2.1.1} and~\ref{fig:2.1.2} are constructed. We first use t-SNE to jointly project the original query embeddings and search momentum from all methods into a two-dimensional space. We then translate the projected points to position the original questions near the origin and center the momentum points at each iteration. During translation, all points within the same group are shifted by the same amount, preserving their relative positions. We then sample trajectories separately for each method and construct three-dimensional plots with the iteration index on the z-axis. For each method at each iteration, we draw a smooth closed contour around the corresponding point cloud to illustrate how search momentum evolves across iterations.

% \subsection{Search Intent and Evidence Alignment across Additional Datasets}
% \label{app:additional-alignment}

% The alignment analysis in Section~\ref{sec:alignment-analysis} focuses on 2WikiQA. Here we provide the corresponding search-intent and query--evidence comparisons for HotpotQA, MuSiQue, and PopQA. Figure~\ref{fig:alignment-app-intent} groups the question--query similarity heatmaps. Figure~\ref{fig:alignment-app-evidence} groups the query--evidence similarity curves.

% \input{image/aligment-app-1}
% \input{image/aligment-app-2}

% As shown in Figure~\ref{fig:alignment-app-intent}, \method{} maintains higher question--query similarity than Search-o1 across HotpotQA, MuSiQue, and PopQA throughout the five iterations, with a more pronounced advantage in later iterations compared with Search-o1. Figure~\ref{fig:alignment-app-evidence} likewise shows stronger and more stable query--evidence alignment for \method{} across these datasets.

\subsection{Case Study}
\label{app:case-study}
Table~\ref{tab:case-study-comparison} and Figure~\ref{fig:case-study-tsne} present a qualitative case study on a 2WikiQA question asking which of \emph{La M\^ome Vert-De-Gris} and \emph{The Pod People} has the earlier-born director. 

Although Search-o1's first retrieval, for ``La M\^ome Vert-De-Gris director,'' identifies Olivier Dahan, its reasoning then invents a version of \emph{The Pod People} directed by Natasha Lyonne and Amy Nicholson. It next queries Amy Nicholson's birth date and drifts further toward documentary production, 1930s organizations, and finally the Farm Holiday Association's membership, following this unsupported interpretation and deviating from the original question . In contrast, \method{}'s parallel queries identify both relevant directors, Bernard Borderie and Juan Piquer Sim\'on, in the first iteration and focus subsequent retrieval on their birth dates. A biographical entry retrieved in the second iteration records Sim\'on's birth on 16 February 1935; later film entries reconfirm Borderie's 1953 directorial debut, from which \method{} infers that he was born earlier, although his exact birth date is not retrieved. This broader initial exploration and evidence-guided refinement enable \method{} to answer correctly without following Search-o1's misleading local trajectory.
\input{image/case_tsne}

\input{table/case_study_pdf}

\subsection{Prompt Templates}
\label{sec:appendix-prompts}

\begin{figure*}[hb]
   \centering
   \includegraphics[width=0.9\linewidth]{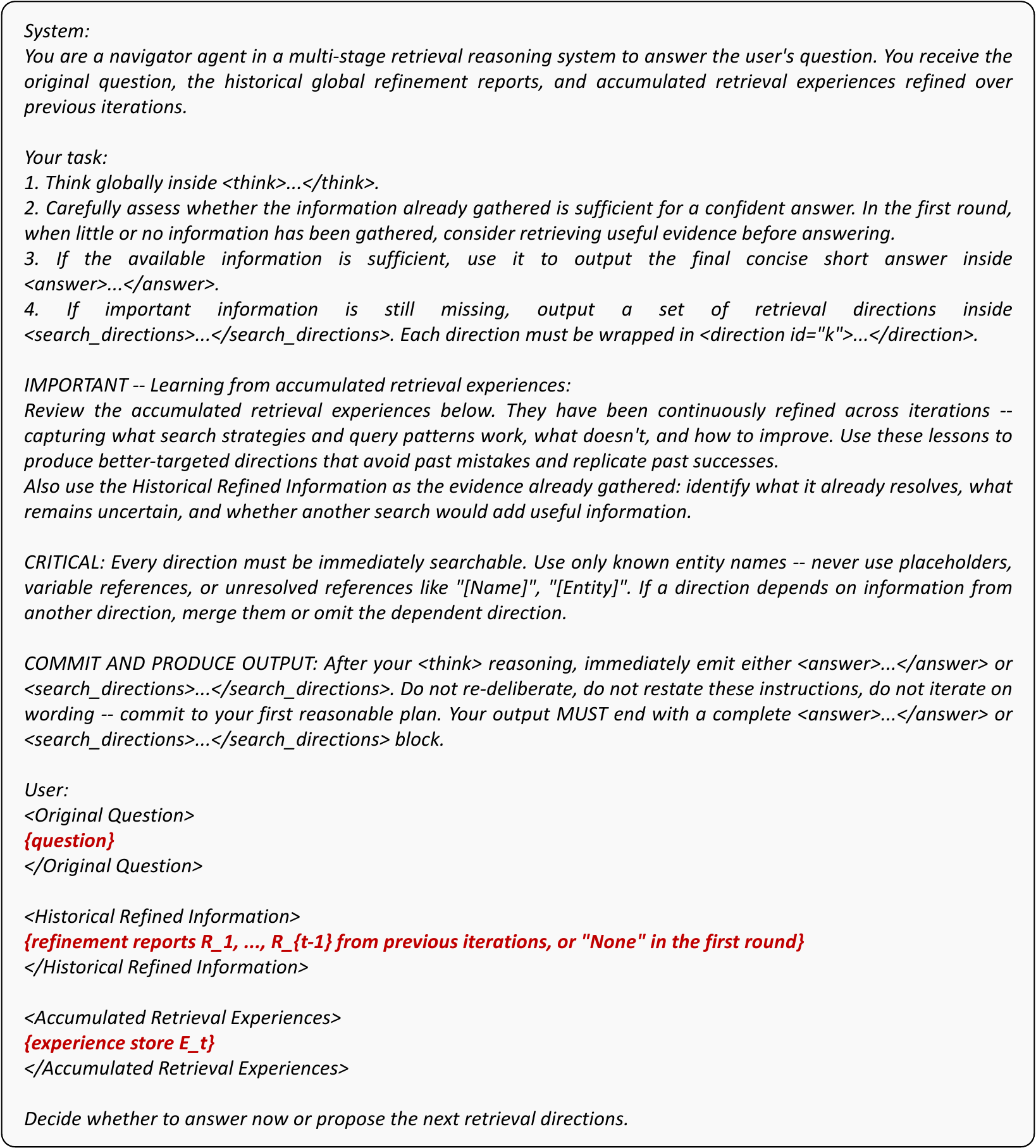}
   \caption{Prompt Template Used in \method{} for Generating Retrieval Directions.}
   \label{fig:prompt-main}
\end{figure*}

\begin{figure*}[hb]
   \centering
   \includegraphics[width=0.9\linewidth]{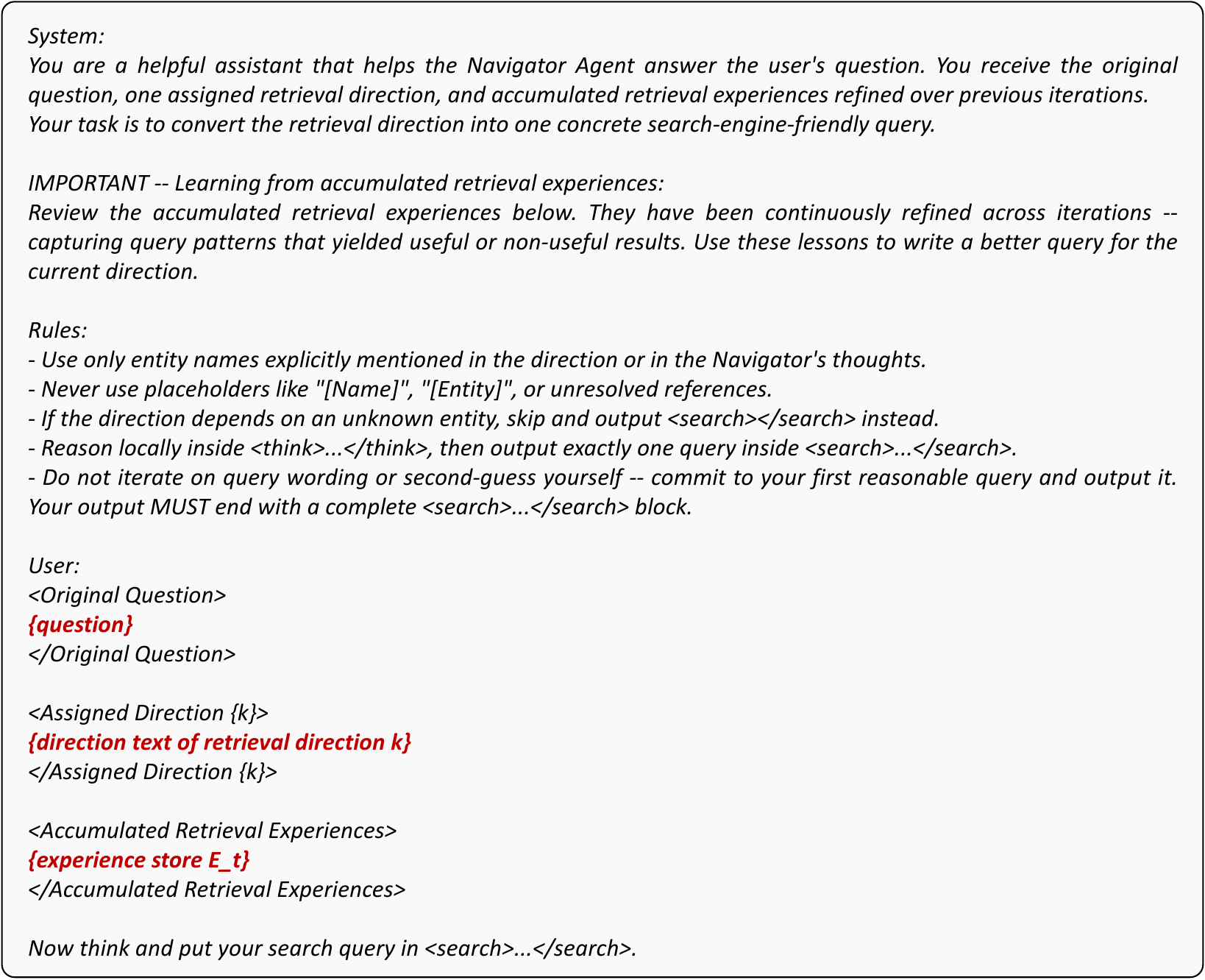}
   \caption{Prompt Template Used in \method{} for Parallel Sub-Query Generation.}
   \label{fig:prompt-parallel}
\end{figure*}

\begin{figure*}[hb]
   \centering
   \includegraphics[width=0.9\linewidth]{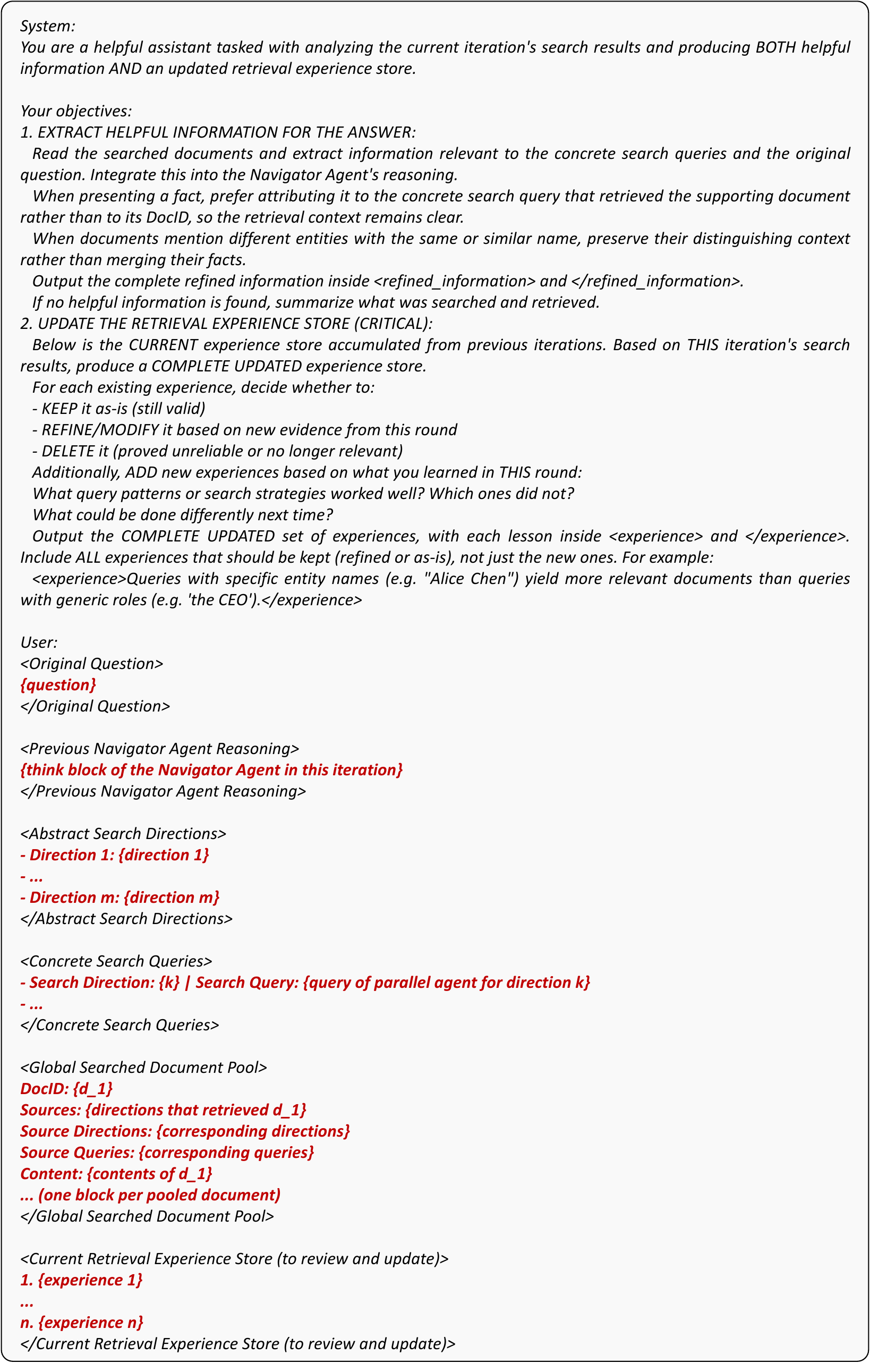}
   \caption{Prompt Template Used in \method{} for Refined Reasoning.}
   \label{fig:prompt-refine}
\end{figure*}

We detail the prompt templates employed in \method{}. As shown in Figure~\ref{fig:prompt-main}, it assesses whether the accumulated knowledge is sufficient to answer the original question and either produces a final answer or proposes complementary retrieval directions. As shown in Figure~\ref{fig:prompt-parallel}, it prompts the LLM to convert these directions into concrete sub-queries for parallel retrieval. Finally, the prompt in Figure~\ref{fig:prompt-refine} extracts relevant information from the retrieved documents and updates the accumulated reasoning traces, providing evidence and guidance for subsequent iterations.

%% file: table/momentum_beta_fit_maps.tex
\begin{table*}[htbp]
  \centering
  \vspace{-1em}
  \caption{Fitted Momentum Coefficients and Directional Alignment between Search Momentum and Subsequent Query Intent Deviations for \method{}.}
  \label{tab:momentum-beta-fit-maps}
  \resizebox{\textwidth}{!}{%
    \begin{tabular}{lccccccccc}
      \toprule
      Dataset & $\beta_1$ & $S_1$ & $\beta_2$ & $S_2$ & $\beta_3$ & $S_3$ & $\beta_4$ & $S_4$ & Avg. \\
      \midrule
      2WikiQA     & 0.0000 & 0.5264 & 0.5408 & 0.8206 & 0.7472 & 0.8689 & 0.8698 & 0.8929 & 0.7772 \\
      MuSiQue     & 0.0000 & 0.5120 & 0.5272 & 0.8297 & 0.7404 & 0.8844 & 0.8394 & 0.9050 & 0.7828 \\
      HotpotQA    & 0.0000 & 0.5851 & 0.5256 & 0.8438 & 0.7259 & 0.9037 & 0.8557 & 0.9131 & 0.8115 \\
      \midrule
      Pooled MHQA & 0.0000 & 0.5331 & 0.5328 & 0.8276 & 0.7408 & 0.8803 & 0.8570 & 0.9006 & 0.7854 \\
      \bottomrule
    \end{tabular}%
  }
\end{table*}

%% file: table/momentum_beta_fit_maps_wo_disentanglement.tex
\begin{table*}[htbp]
  \centering
  \caption{Fitted Momentum Coefficients and Held-out Alignment Scores for MAPS w/o Disentanglement on Three Multi-hop Datasets and Pooled MHQA.}
  \label{tab:momentum-beta-fit-maps-ablation}
  \resizebox{\textwidth}{!}{%
    \begin{tabular}{lccccccccc}
      \toprule
      Dataset & $\beta_1$ & $S_1$ & $\beta_2$ & $S_2$ & $\beta_3$ & $S_3$ & $\beta_4$ & $S_4$ & Avg. \\
      \midrule
      2WikiQA     & 0.0000 & 0.5036 & 0.5469 & 0.8102 & 0.7543 & 0.8548 & 0.8527 & 0.8533 & 0.7555 \\
      MuSiQue     & 0.0000 & 0.4855 & 0.5369 & 0.7824 & 0.7538 & 0.8329 & 0.8629 & 0.8545 & 0.7388 \\
      HotpotQA    & 0.0000 & 0.5435 & 0.5467 & 0.8086 & 0.7607 & 0.8236 & 0.8651 & 0.8596 & 0.7588 \\
      \midrule
      Pooled MHQA & 0.0000 & 0.5065 & 0.5422 & 0.7978 & 0.7549 & 0.8396 & 0.8589 & 0.8547 & 0.7497 \\
      \bottomrule
    \end{tabular}%
  }
\end{table*}

%% file: image/case_tsne.tex
% \begin{figure}[tb]
%   \centering
%   \vspace{-1em}
%   \subfigure[Document Embedding Space]{
%     \label{fig:case-tsne-1}
%     \includegraphics[width=0.32\textwidth]{fig/case_study/query_evolution_all_6/query_evolution_all_6_1.pdf}}\hfill
%   \subfigure[Iteration 1]{
%     \label{fig:case-tsne-2}
%     \includegraphics[width=0.32\textwidth]{fig/case_study/query_evolution_all_6/query_evolution_all_6_2.pdf}}\hfill
%   \subfigure[Iteration 2]{
%     \label{fig:case-tsne-3}
%     \includegraphics[width=0.32\textwidth]{fig/case_study/query_evolution_all_6/query_evolution_all_6_3.pdf}}\hfill
%   \subfigure[Iteration 3]{
%     \label{fig:case-tsne-4}
%     \includegraphics[width=0.32\textwidth]{fig/case_study/query_evolution_all_6/query_evolution_all_6_4.pdf}}\hfill
%   \subfigure[Iteration 4]{
%     \label{fig:case-tsne-5}
%     \includegraphics[width=0.32\textwidth]{fig/case_study/query_evolution_all_6/query_evolution_all_6_5.pdf}}\hfill
%   \subfigure[Iteration 5]{
%     \label{fig:case-tsne-6}
%     \includegraphics[width=0.32\textwidth]{fig/case_study/query_evolution_all_6/query_evolution_all_6_6.pdf}}\hfill
%   \caption{T-SNE Projection of E5 Embeddings for the Case-study Trajectories in Table~\ref{tab:case-study-comparison}. The black diamond marks the original question, and gold stars mark the four gold documents. Orange points trace Search-o1's sequential queries; blue points mark \method{}'s parallel queries, with letters identifying branches. Blue outlines indicate the range covered by \method{}'s queries in each iteration.}
%     \label{fig:case-study-tsne}
%   \end{figure}

\begin{figure}[tb]
  \centering
  \vspace{-1em}

  \begin{subfigure}[t]{0.32\textwidth}
    \centering
    \includegraphics[width=\linewidth]{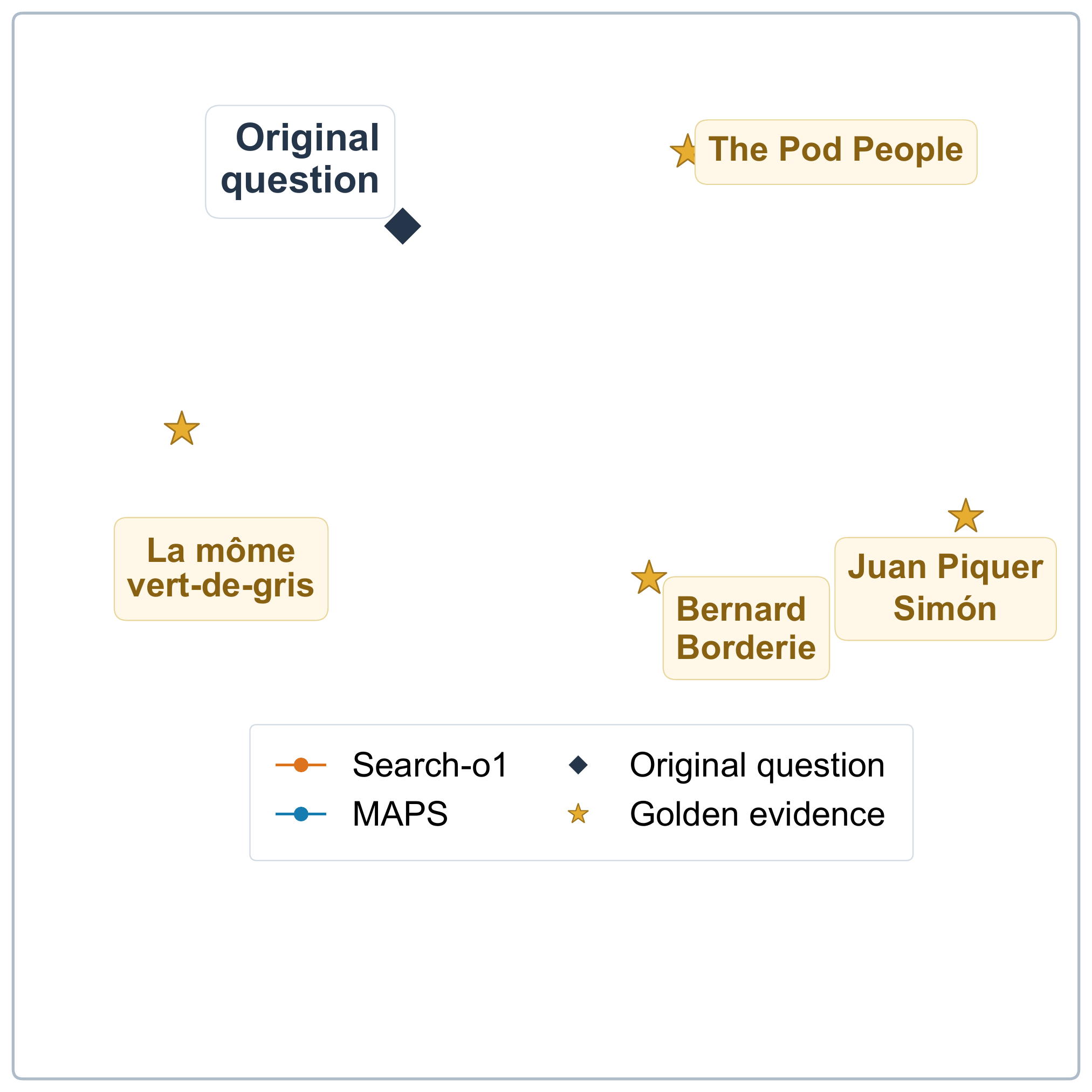}
    \caption{Document Embedding Space}
    \label{fig:case-tsne-1}
  \end{subfigure}\hfill
  \begin{subfigure}[t]{0.32\textwidth}
    \centering
    \includegraphics[width=\linewidth]{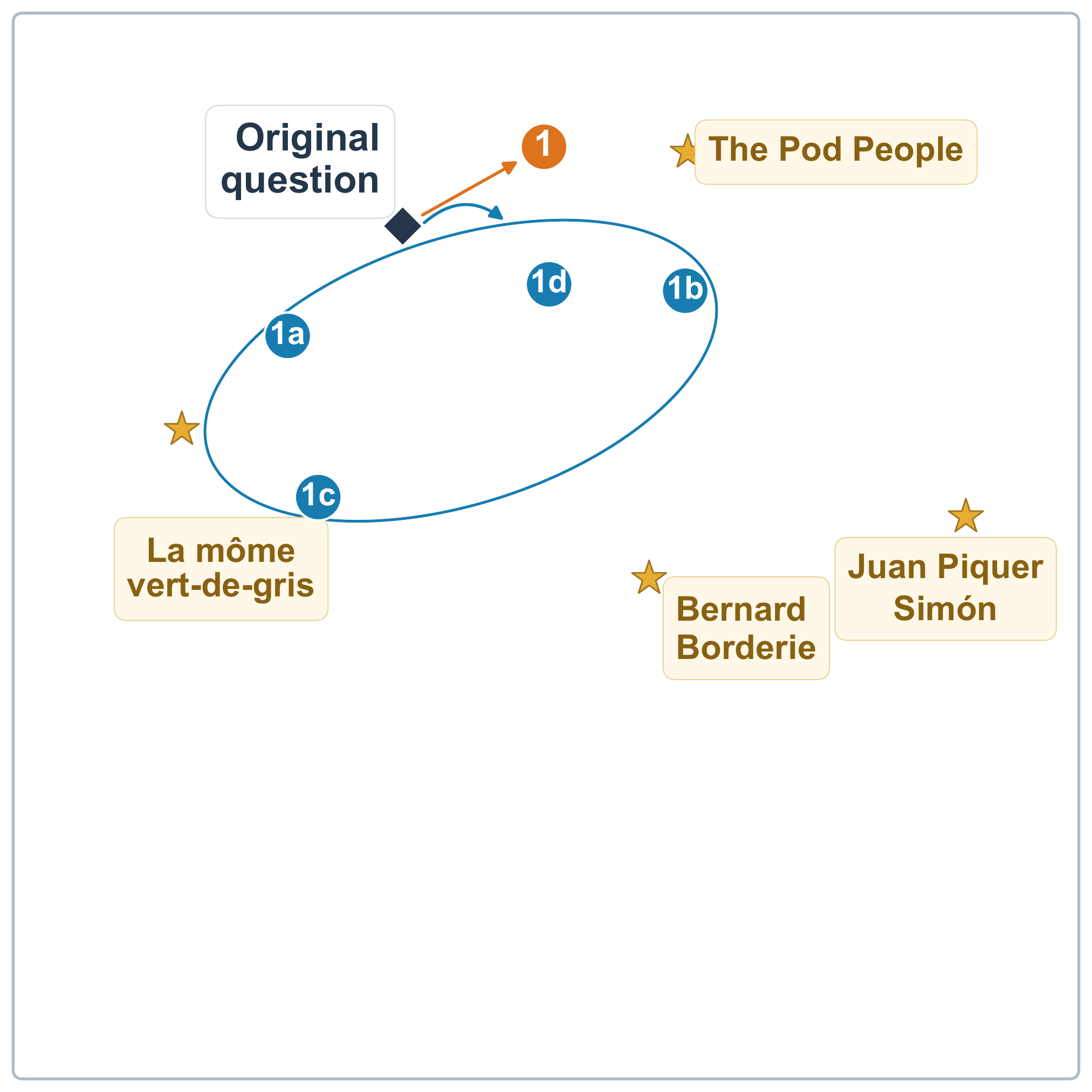}
    \caption{Iteration 1}
    \label{fig:case-tsne-2}
  \end{subfigure}\hfill
  \begin{subfigure}[t]{0.32\textwidth}
    \centering
    \includegraphics[width=\linewidth]{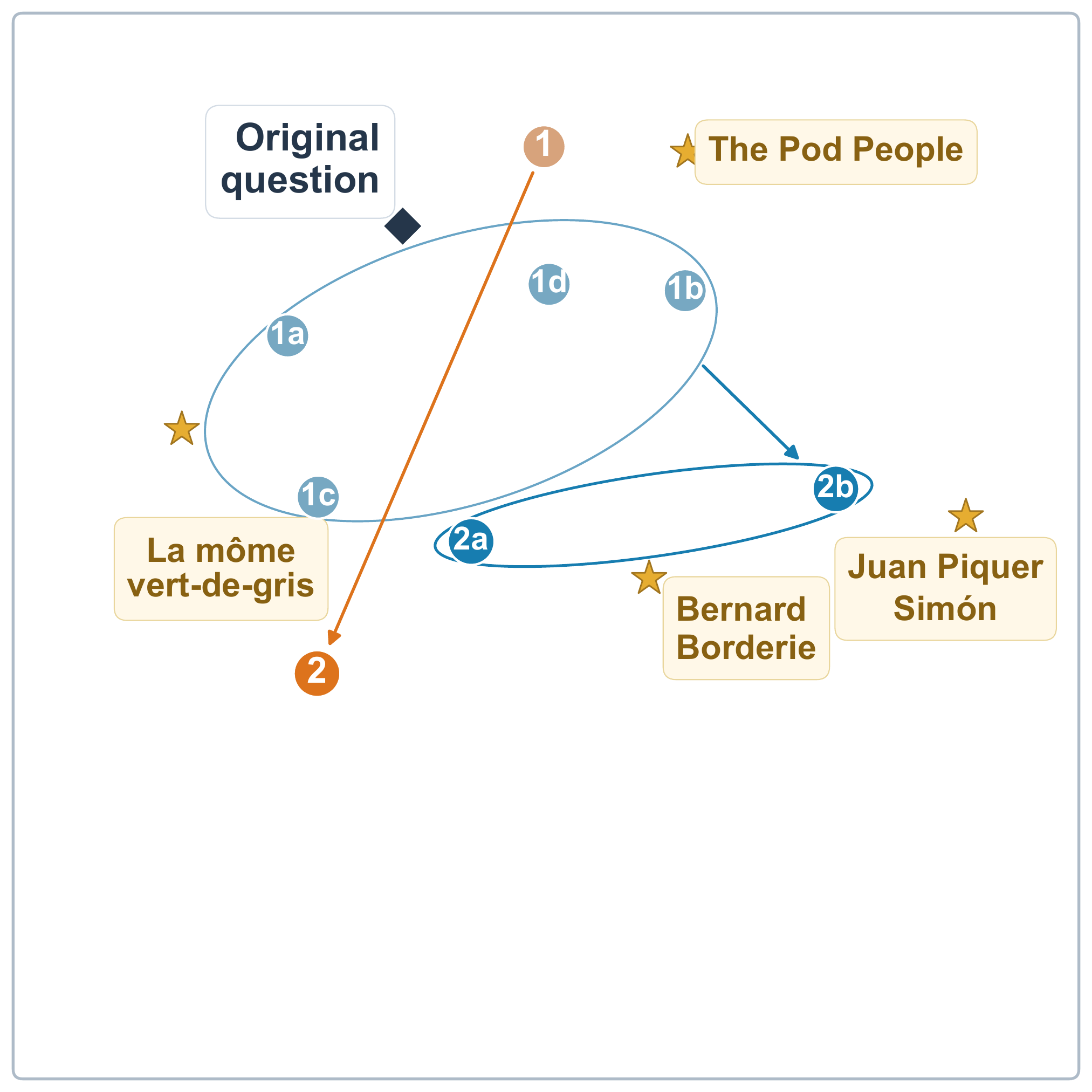}
    \caption{Iteration 2}
    \label{fig:case-tsne-3}
  \end{subfigure}\hfill

  \begin{subfigure}[t]{0.32\textwidth}
    \centering
    \includegraphics[width=\linewidth]{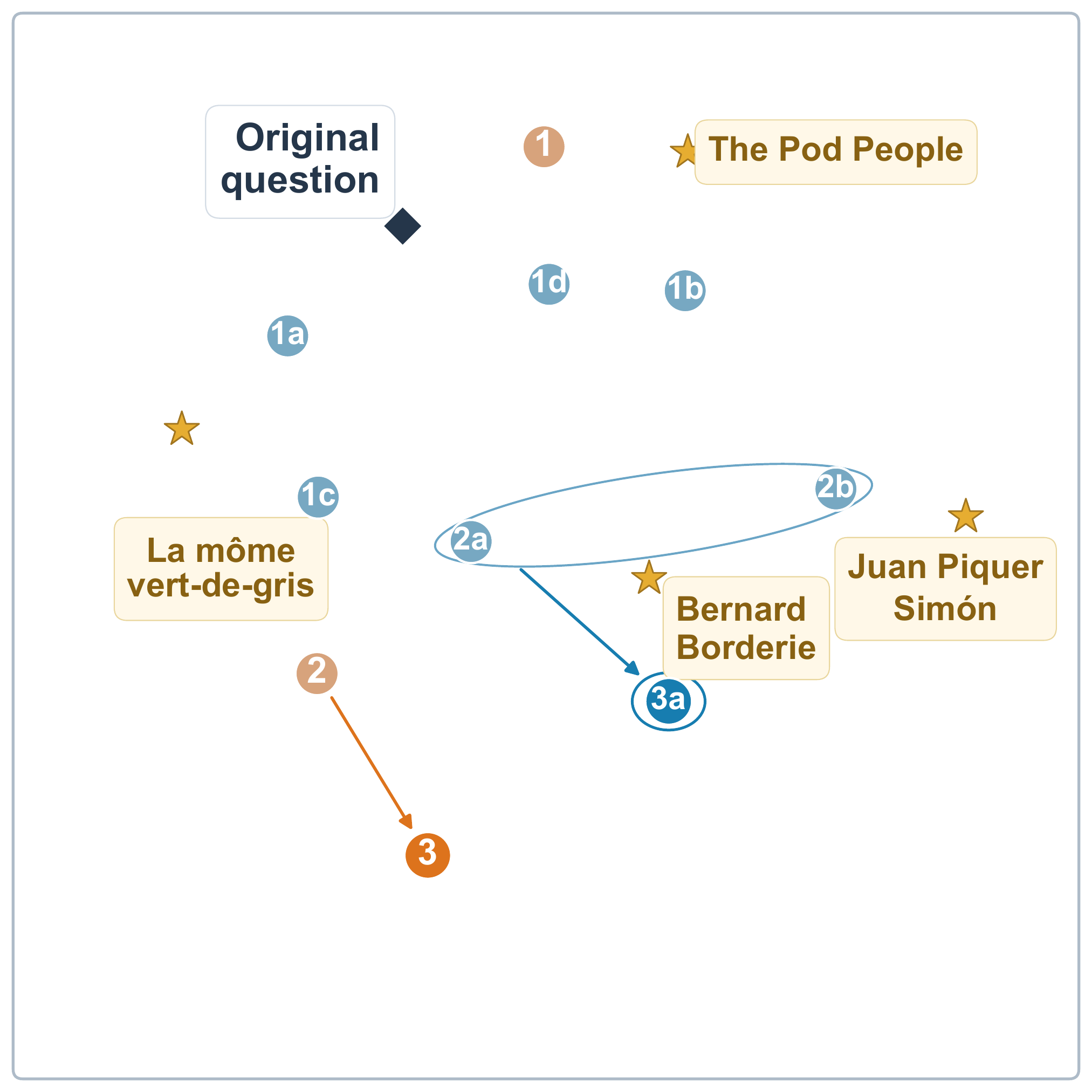}
    \caption{Iteration 3}
    \label{fig:case-tsne-4}
  \end{subfigure}\hfill
  \begin{subfigure}[t]{0.32\textwidth}
    \centering
    \includegraphics[width=\linewidth]{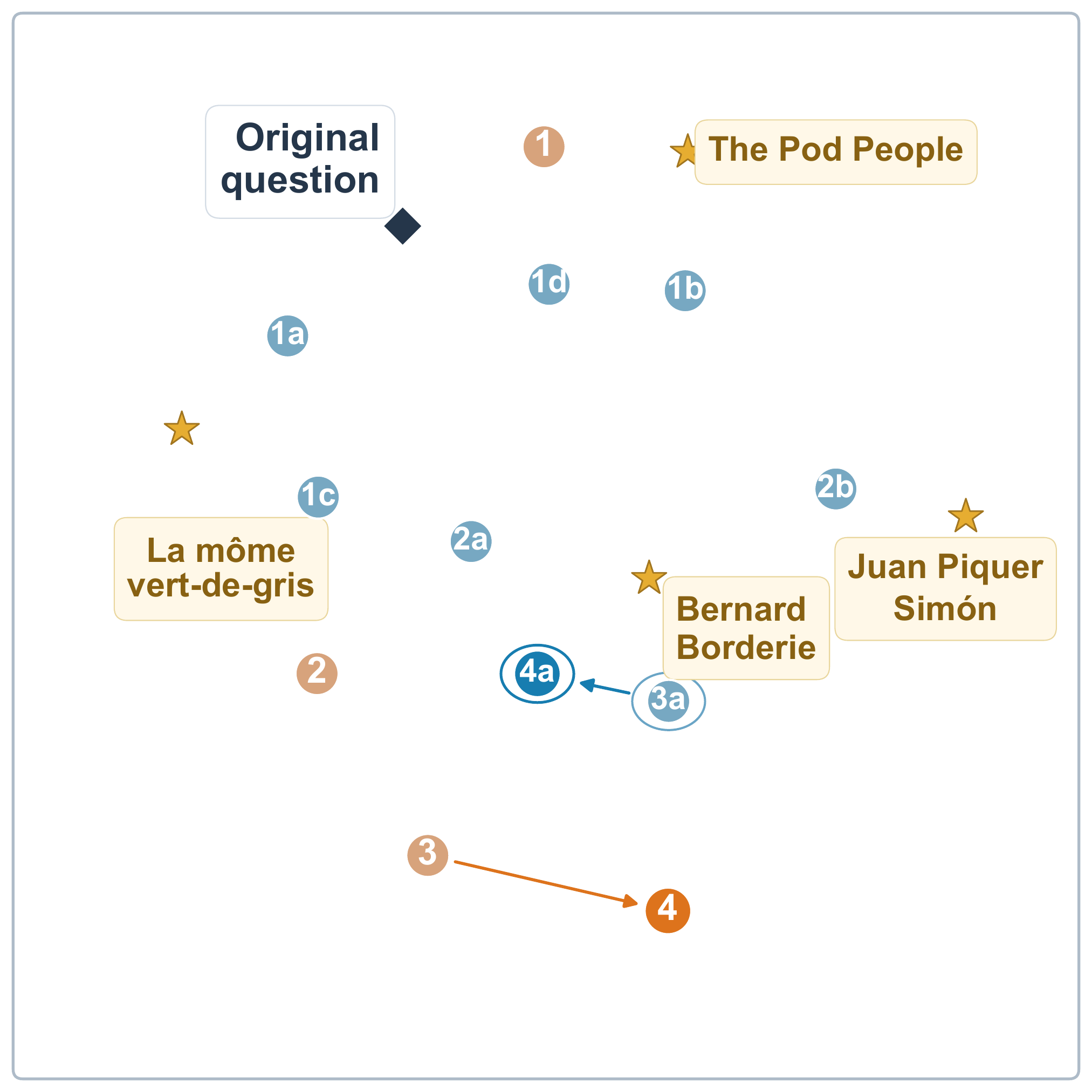}
    \caption{Iteration 4}
    \label{fig:case-tsne-5}
  \end{subfigure}\hfill
  \begin{subfigure}[t]{0.32\textwidth}
    \centering
    \includegraphics[width=\linewidth]{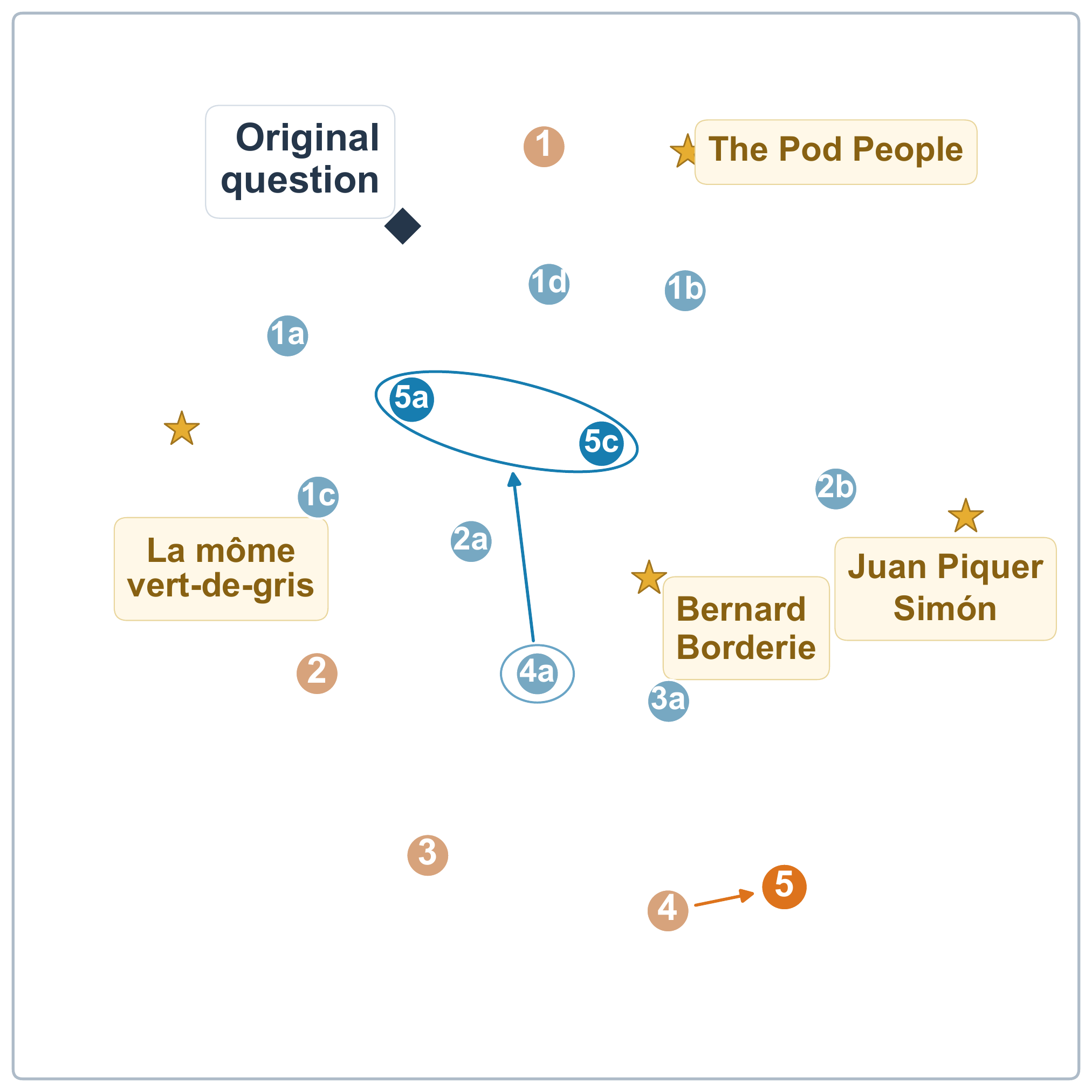}
    \caption{Iteration 5}
    \label{fig:case-tsne-6}
  \end{subfigure}

  \caption{T-SNE Projection of E5 Embeddings for the Case-study Trajectories in Table~\ref{tab:case-study-comparison}. The black diamond marks the original question, and gold stars mark the four gold documents. Orange points trace Search-o1's sequential queries; blue points mark \method{}'s parallel queries, with letters identifying branches. Blue outlines indicate the range covered by \method{}'s queries in each iteration.}
  \label{fig:case-study-tsne}
\end{figure}

%% file: table/case_study_pdf.tex
% The vector PDF is compiled from fig/case_study/case_study.tex.
\begin{table*}[t]
  \centering
  \caption{Case Study Comparing Search-o1 and \method{}.  Red highlights misleading observations, off-topic evidence associated with query drift, and the incorrect final output; blue marks correctly identified key entities and the correct final answer.}
  \label{tab:case-study-comparison}
  \label{tab:case-study-search-o1}
  \label{tab:case-study-maps}
  \includegraphics[width=\linewidth]{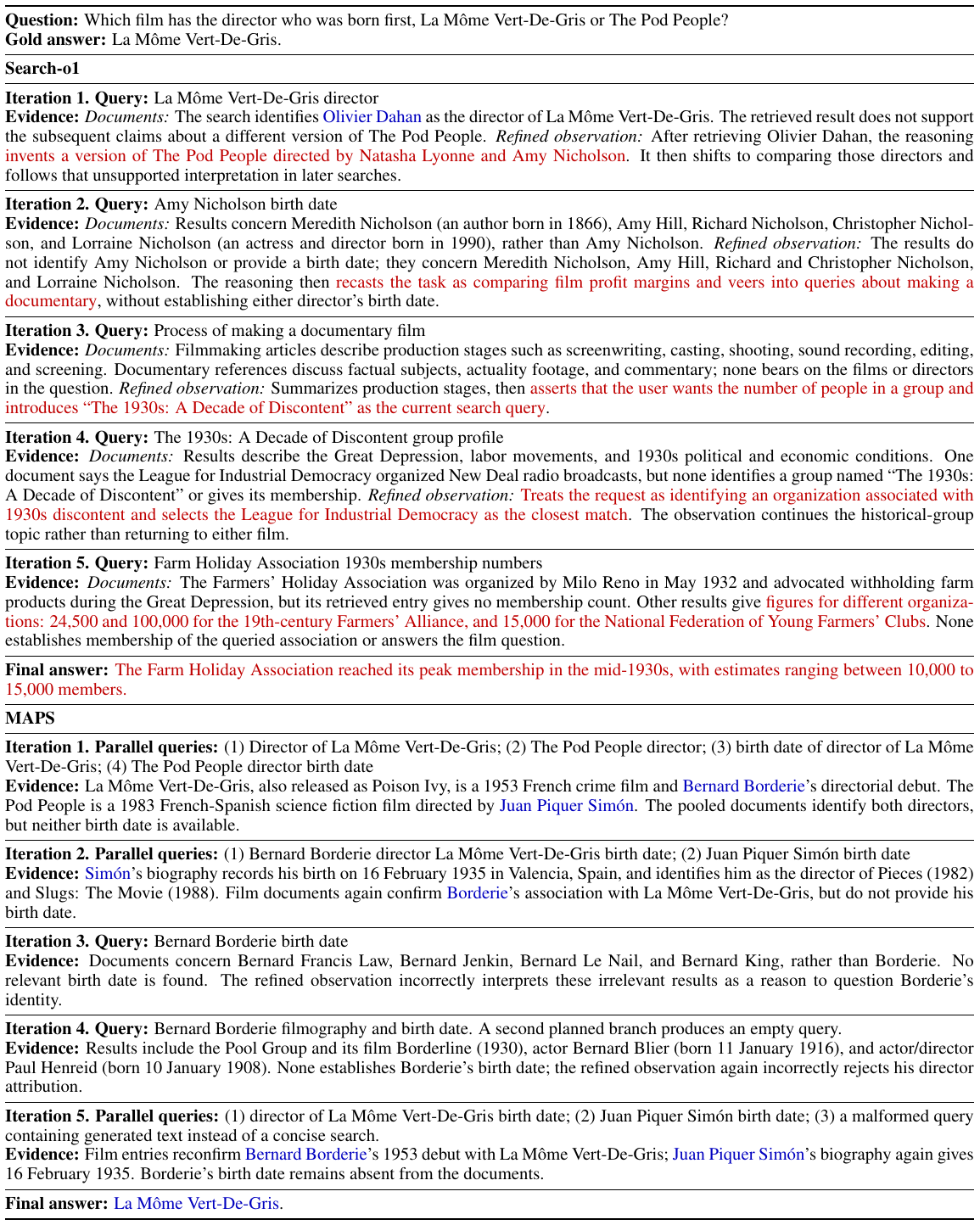}
  \vspace{-1em}
\end{table*}